\documentclass[journal]{IEEEtran}
\makeatletter

\newcommand{\Rnum}[1]{\expandafter\@slowromancap\romannumeral #1@}
\usepackage{eufrak}
\usepackage{hyperref}
\usepackage{ifpdf}
\usepackage{multirow}
\usepackage{cite}
\ifCLASSINFOpdf
   \usepackage[pdftex]{graphicx}
   \graphicspath{{../pdf/}{../jpeg/}}
   \DeclareGraphicsExtensions{.pdf,.jpeg,.png}
\else
   \usepackage[dvips]{graphicx}
   \graphicspath{{../eps/}}
   \DeclareGraphicsExtensions{.eps}
\fi
\usepackage{amsmath}
\usepackage{amssymb}
\usepackage{amsfonts}
\usepackage{algorithmic}
\usepackage{array}
\ifCLASSOPTIONcompsoc
  \usepackage[caption=false,font=normalsize,labelfont=sf,textfont=sf]{subfig}
\else
  \usepackage[caption=false,font=footnotesize]{subfig}
\fi
\usepackage{url}
\usepackage{color}
\usepackage{tablefootnote}
\begin{document}

\title{
TOAN: Target-Oriented Alignment Network 
for Fine-Grained Image Categorization \\ with Few Labeled Samples}

\author{Huaxi Huang,
        Junjie Zhang,
        Litao Yu,
        Jian Zhang,~\IEEEmembership{Senior Member,~IEEE,}\\
        Qiang Wu,~\IEEEmembership{Senior Member,~IEEE,}
        Chang Xu.

\thanks{
Copyright © 2021 IEEE. Personal use of this material is permitted. However, permission to use this material for any other purposes must be obtained from the IEEE by sending an email to pubs-permissions@ieee.org.

Corresponding author: Jian Zhang, email: Jian.Zhang@uts.edu.au. 
Huaxi Huang and Junjie Zhang are co-first authors.

Huaxi Huang, Litao Yu, Jian Zhang and Qiang Wu are with the Faculty of Engineering and Information Technology, University of Technology Sydney, Sydney NSW 2007, Australia.
Junjie Zhang is with the Key Laboratory of Specialty Fiber Optics and Optical Access Networks, Joint International Research Laboratory of Specialty Fiber Optics and Advanced Communication, Shanghai Institute of Advanced Communication and Data Science, Shanghai University, Shanghai 200444, China.
Chang Xu is with the School of Computer Science, The University of Sydney, Sydney NSW 2006, Australia.
}
}

\markboth{Journal of \LaTeX\ Class Files,~Vol.~XX, No.~X, June~2019}%
{Shell \MakeLowercase{\textit{et al.}}: Bare Demo of IEEEtran.cls for IEEE Journals}

\maketitle

\begin{abstract}

\textcolor{black}{In this paper, we study the fine-grained categorization problem under the few-shot setting, i.e., each fine-grained class only contains a few labeled examples, termed Fine-Grained Few-Shot classification (FGFS). 
The core predicament in FGFS is the high intra-class variance yet low inter-class fluctuations in the dataset. 
In traditional fine-grained classification, the high intra-class variance can be somewhat relieved by conducting the supervised training on the abundant labeled samples. However, with few labeled examples, it is hard for the FGFS model to learn a robust class representation with the significantly higher intra-class variance. Moreover, the inter- and intra-class variance are closely related. The significant intra-class variance in FGFS often aggravates the low inter-class variance issue.}

\textcolor{black}{To address the above challenges, we propose a Target-Oriented Alignment Network (TOAN) to tackle the FGFS problem from both intra- and inter-class perspective.
To reduce the intra-class variance, we propose a target-oriented matching mechanism to reformulate the spatial features of each support image to match the query ones in the embedding space.
To enhance the inter-class discrimination, we devise discriminative fine-grained features by integrating local compositional concept representations with the global second-order pooling.
We conducted extensive experiments on four public datasets for fine-grained categorization, and the results show the proposed TOAN obtains the state-of-the-art.
}

\end{abstract}


\IEEEpeerreviewmaketitle

\section{Introduction}
\IEEEPARstart{F}{ine-Grained}
(FG) visual recognition aims to distinguish different sub-categories belonging to the same entry-level category, such as animal identification~\cite{KhoslaYaoJayadevaprakashFeiFei_FGVC2011,Horn_2015_CVPR,WahCUB_200_2011} and vehicle recognition~\cite{KrauseStarkDengFei-Fei_3DRR2013}. 
Existing FG models~\cite{Cui_2017_CVPR,Fu_2017_CVPR,Krause_2015_CVPR,Li_2018_CVPR,Lin_2015_ICCV,Yu_2018_ECCV,zhang2014part,zhang2018fine,zheng2019learning,wang2019deep,Han} utilize large-scale and fully-annotated training sets to `understand' and `memorize' the \textcolor{black}{subtle differences among classes,} thus achieving satisfactory performances in identifying new samples from the same label space. 
However, in many practical scenarios, it is hard to obtain abundant labeled data for fine-grained classification. For example, in industrial defects detection, most defects exist only in a few common categories, while most other categories only contain a small portion of defects. 
Moreover, annotating a large-scale fine-grained dataset is labor-intensive, which requires high expertise in many fields. Thus, how to obtain an effective model with sparse labeled samples remains an open problem.
In this paper, we focus on one of the limited sample learning methods for FG tasks, \textit{i.e.,} Fine-Grained image classification under Few-Shot settings (FGFS).

\begin{figure}[t]
\centering
\includegraphics[width=1.0\linewidth]{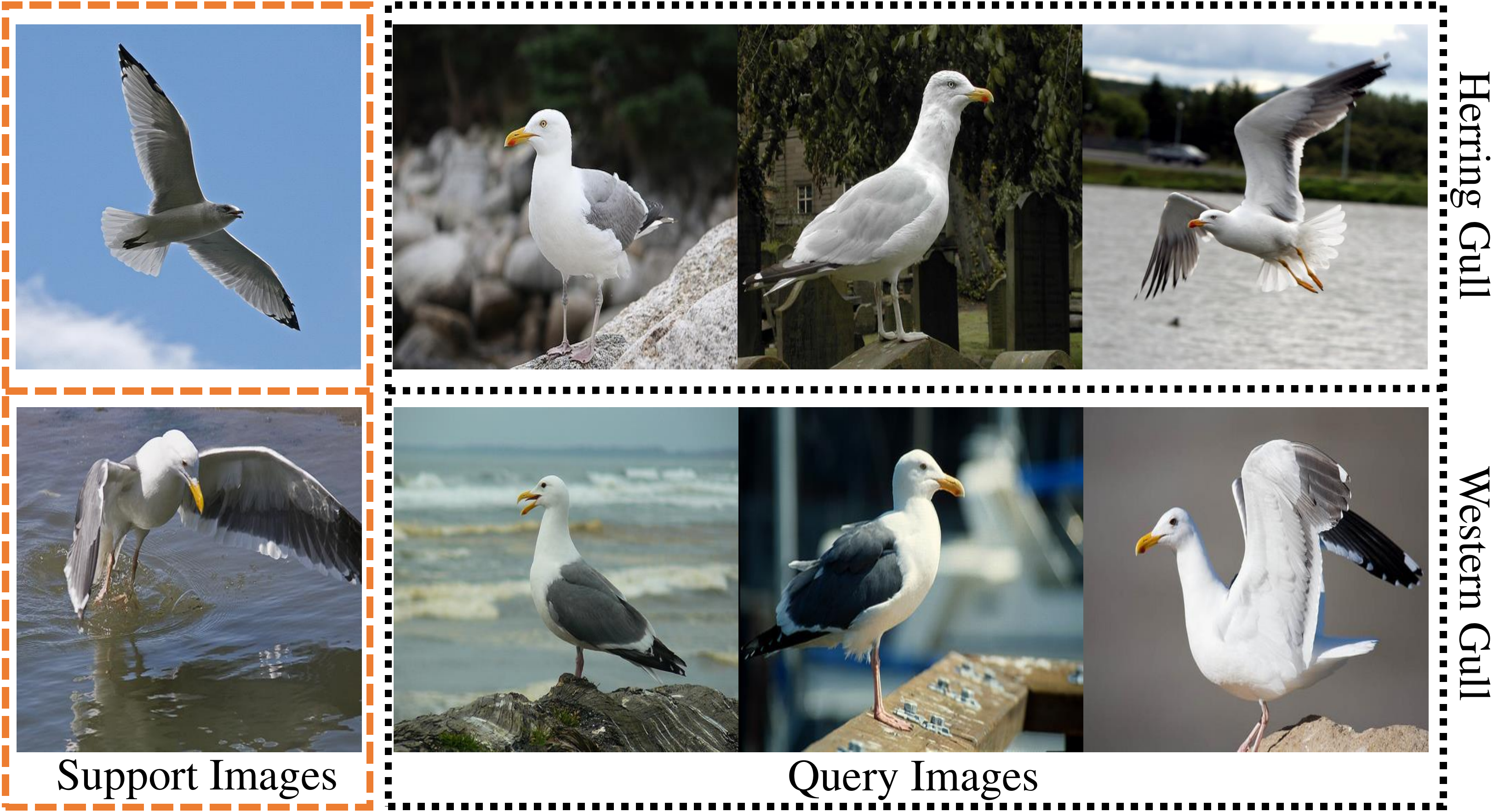}
\caption{The high inter-class visual similarity and significant intra-class variations in FGFS tasks are more rigorous than general FG tasks. Some Herring gull and western gull images have similar visual appearances, which indicates the subtle inter-class variance.
However, in each class, gulls present different postures with different backgrounds, which brings significant intra-class variance.}
\label{bird}
\end{figure}

\begin{figure*}[t]
\begin{center}
\includegraphics[width=1.0\linewidth]{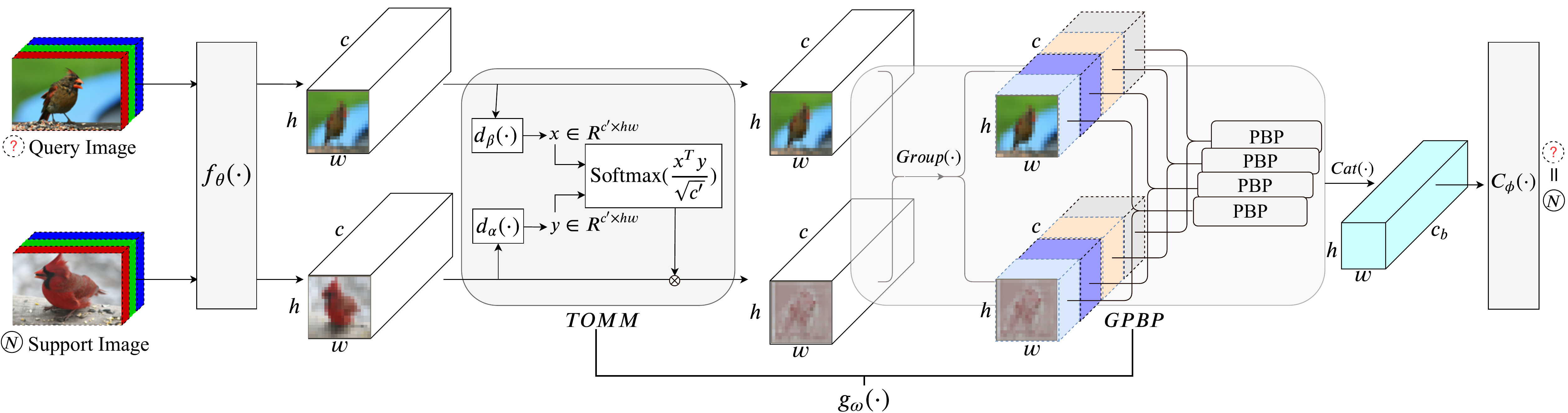}
\end{center}
   \caption{\textcolor{black}{The overview of proposed TOAN in the N-way-1-shot FGFS task, other support samples are omitted (replaced by $N$). The model consists of three parts: the feature embedding $f_{\theta}$ learns the convolved features;
   the fine-grained relation extractor $g_\omega$ contains target-orientated matching mechanism (TOMM) and group pair-wise bilinear pooling (GPBP), TOMM aims at reformulating the features of support image to match the query image feature in the embedding space through the cross-correction attention mechanism, while GPBP is designed to extract discriminative second-order features by incorporating the channel grouping. With TOMM and GPBP, $g_\omega$ learns to generate robust bilinear features from support-query pairs. PBP stands for the Pair-wise Bilinear Pooling,
   and the comparator $C_\phi$ maps the query to its ground-truth class.}
   }
\label{framework}
\end{figure*}

\textcolor{black}{The core challenges of the FG problem are the high intra-class variance and low inter-class fluctuations within the datasets~\cite{Fu_2017_CVPR,Lin_2015_ICCV}. 
The high intra-class variance is mainly caused by different viewpoints, spatial poses, motions, and lighting conditions of different samples in each class. On the other hand, the subtle inter-class variance reflects the taxonomy definition that different fine-grained categories belong to the same entry-level category.
The ideal data distribution in a classification problem should posses the low intra-class variance with the high inter-class variance.
If the low inter-class variance is accompanied by high intra-class variance, it can easily lead to inaccurate classification boundaries.
With large-scale and fully-annotated datasets available, the high intra-class variance can be somehow relieved through supervised training to obtain a robust representation of each class. 
However, for FGFS, each class only contains limited labeled samples. As is seen from Fig.~\ref{bird}, in the one-shot bird classification scenario, if the single support (labeled) sample shows a diving gesture, while query (unlabeled) ones are standing. The query-support pairs are not spatially aligned, which can be `confusing' for classifiers to distinguish them. Therefore, the large intra-class differences bring significant impacts on the representation learning in FGFS.}
Unfortunately, current FGFS models~\cite{li2019CovaMNet,wei2019piecewise,wertheimer2019few,zhang2019power} rarely focus on this issue. \textcolor{black}{To solve the query-support unmatched issue, we propose to adopt the cross-correction attention to transfer the support image features to spatially align with the query ones.}
From the perspective of feature representation, the second-order representation learning~\cite{huang2019low,li2019CovaMNet,wei2019piecewise,wertheimer2019few,zhang2019power} is usually carried out to address the low inter-class variance in FGFS tasks. They focus on enhancing FG features from the global view. 
The global features describe the overall appearances of image instances but fail to describe the local visual properties, which is less discriminative in FGFS classification tasks.
\textcolor{black}{To improve the discrimination of the second-order features, we propose to mine the local concept features using a group pair-wise bilinear pooling operation.}
\textcolor{black}{
We name the whole model as \textbf{Target-Oriented Alignment Network} (TOAN).
By jointly employing the feature alignment transformation to reduce the high intra-class variance and the second-order comparative feature extraction to enlarge the low inter-class discrimination, we explore robust fine-grained relations between each support-query pair. The whole framework is shown in Fig. \ref{framework}.}

\textcolor{black}{More specifically, to address the high intra-class variance with limited supervision, we propose a Target-Oriented Matching Mechanism (TOMM), which is inspired by the classical template-based fine-grained methods.}
As a decent solution to alleviate the high intra-class variance in traditional fine-grained classification, template-based fine-grained methods \cite{tmp1,tmp2,tmp3,tmp4} utilize the templates (the closest samples or parts to the class centroid) to align the samples in each class. 
However, in FGFS, the labeled samples in each class are extremely limited. It is hard to select a good template to represent each category. 
Therefore, we set each query (testing) sample as the template for all the support (labeled) samples and then adopt the cross-correction attention to transform support features to match query ones.
TOMM reformulates the convolutional representations of support images by comparing them with the spatial features of the target query. 
Different from the conventional self-attention mechanism~\cite{vaswani2017attention} that operates on the input feature itself, we propose to generate the attention weights in a target-oriented fashion.
That is, the similarities of the convolutional features between the support-query pairs are computed first and then converted into a soft-attention map, also noted as the cross-correction attention~\cite{hou2019cross}. The spatial features of the support image are then recomputed as the weighted sum of the whole feature map to reduce the possible variance compared to the query.

To address the low inter-class variance, different from existing works focus on devising high order features from the global view, we propose to mine the concept local compositionality representation of the bilinear features to enhances their discriminate ability.
Compositionality helps humans learn new concepts from limited samples since it can convert concepts to knowing primitive~\cite{biederman1987recognition,hoffman1984parts,marr1978representation}. 
For a convolutional neural network, the channels of convolutional feature usually correspond to different sets of visual patterns~\cite{simon2015neural,zhang2016picking,zheng2017learning}. Therefore, inspired by \cite{hu2019weakly,zhang2017interleaved,zheng2017learning,zheng2019learning}, we incorporate the compositional concepts into the fine-grained feature extraction by combining the channel grouping operation with the pair-wise bilinear pooling, noted as Group Pair-wise Bilinear Pooling (GPBP).

In summary, this paper makes the following contributions:
\begin{itemize}
    \item We propose a Target-Oriented Matching Mechanism (TOMM) to learn explicit feature transformations to reduce the biases caused by the intra-class variance. By adopting a cross-correction attention mechanism, the target-oriented matching transfers the support image features to spatially align with the query ones.
    \item  We propose to aggregate the regional representations into pairwise bilinear pooling through the convolutional channel grouping (GPBP), which devises the second-order features from both global and local views. To our best knowledge, this is the first attempt to adopt group bilinear pooling in FGFS. 
    \item We conducted comprehensive experiments on four fine-grained benchmark datasets to investigate the effectiveness of the proposed model, and our model achieves the state-of-the-arts.
\end{itemize}

The rest of this paper is organized as follows. Section \ref{RL} introduces related works. Section \ref{Method} presents the proposed TOAN method. In Section \ref{exp}, we evaluate the proposed method on four widely-used fine-grained datasets. The conclusion is discussed in Section \ref{con}.

\section{Related Work} \label{RL}

\subsection{Fine-Grained Image Categorization}
Fine-grained image classification is a trending topic in the computer vision community in recent years.
Most fine-grained models~\cite{Han,Chen_2019_CVPR,engin2018deepkspd,Fu_2017_CVPR,Ge_2019_CVPR,Krause_2015_CVPR,Li_2018_CVPR,Lin_2015_ICCV,Yu_2018_ECCV,zhang2014part,zhang2016picking,zheng2017learning,zheng2019looking,he2019fine} can be roughly grouped into two categories: regional feature-based models~\cite{Han,Chen_2019_CVPR,Fu_2017_CVPR,Ge_2019_CVPR,Krause_2015_CVPR,zhang2016picking,zheng2017learning} and global feature-based methods~\cite{cai2017higher,Cui_2017_CVPR,engin2018deepkspd,koniusz2018deeper,Li_2018_CVPR,Lin_2015_ICCV,Yu_2018_ECCV,zheng2019learning,hu2020attentional}.
For fine-grained images, the most informative features generally lie in the discriminate parts of the object. Therefore, regional feature-based approaches tend to mine these parts first and then fuse them to form a robust representation of the object. 
\textcolor{black}{
For instance, \cite{Han} proposed a P-CNN model to learn the distinctive parts of the fine-grained images using a part localization network. The extracted parts and the global feature are then jointly fed into the part classification network to generate the final prediction.
In \cite{Chen_2019_CVPR}, the authors adopted the adversarial learning by `constructing' and `destructing' the input image to integrate discriminative features, while the attention mechanisms~\cite{Fu_2017_CVPR,zheng2017learning,zheng2019looking} are employed to remove the non-discriminative parts automatically. 
}

On the other hand, global feature-based fine-grained methods~\cite{cai2017higher,Cui_2017_CVPR,engin2018deepkspd,koniusz2018deeper,Li_2018_CVPR,Lin_2015_ICCV,Yu_2018_ECCV,zheng2019learning,hu2020attentional} extract the feature from the whole image without explicitly localizing the object parts.
Bilinear CNN model (BCNN)~\cite{Lin_2015_ICCV} is the first work that adopts matrix outer product operation on the embedding to generate a second-order representation for fine-grained classification. Li \textit{et al.}~\cite{Li_2018_CVPR} (iSQRT-COV) further improved the naive bilinear model by using covariance matrices over the last convolutional features as fine-grained features. iSQRT-COV obtains  the state-of-art performance on both generic and fine-grained datasets.
Gao \textit{et al.}~\cite{Yu_2018_ECCV} devised a hierarchical approach by employing a cross-layer factorized bilinear pooling operation. 
However, the feature dimensions of the second-order models are the square fold of the naive ones. To reduce the computation complexity,  \cite{Gao_2016_CVPR,Kong_2017_CVPR,kim2016hadamard,Yu_2018_ECCV} adopted the low-rank approximation strategies. For instance, Kim \textit{et al.} \cite{kim2016hadamard} used the Hadamard product to redefine the bilinear matrix outer product and proposes a factorized low-rank bilinear pooling for multimodal learning. Gao \textit{et al.}~\cite{Yu_2018_ECCV} devised a hierarchical approach by employing a cross-layer factorized bilinear pooling operation. 
Our work is inspired by global feature-based methods to generate second-order features for FGFS tasks.

\subsection{Fine-Grained Categorization with Few Labeled Samples}
Wei \textit{et al.} \cite{wei2019piecewise} proposed a Piecewise Classifier Mappings (PCM) framework for fine-grained image categorization under the few-shot setting. PCM injects the bilinear feature~\cite{Lin_2015_ICCV} into a group of mapping networks to reduce the dimensionality of the features. A deep distance classifier is then appended to generate the final prediction.
SoSN~\cite{zhang2019power} adopts the power normalizing second-order pooling to generate the fine-grained features, and a pair-wise mechanism is then proposed to capture the correlation of support-query pairs.
Li \textit{et al.}~\cite{li2019CovaMNet} replaced the bilinear pooling with a covariance pooling operation, and a covariance metric is proposed as the distance classifier. 
Moreover, \cite{wertheimer2019few} designs a localization network to generate the foreground and background features for an input image with external bounding box annotations. The bilinear-pooled foreground and background features are concatenated and fed into the classifier. 
In \cite{li2019CovaMNet,wei2019piecewise,wertheimer2019few,zhang2019power}, the authors adopted the second-order pooling on the input image itself (noted as self-bilinear pooling) to capture the fine-grained representation.
To further leverage the second-order pairwise relationship between support and query images, our previous works propose the pairwise bilinear pooling~\cite{huang2019compare,huang2019low}, of which, \cite{huang2019compare} adopts the matrix-outer-product pooling to model pairwise relationships, and \cite{huang2019low} proposes a factorized Hadamard-product low-rank bilinear operation.
However, the high intra-class variance issue is not explicitly addressed in these works ~\cite{wei2019piecewise,zhang2019power,li2019CovaMNet,wertheimer2019few,huang2019compare}.

It is worth noting that our previous work \cite{huang2019low} presents a feature position re-arrangement module for feature alignment with a global MSE loss to boost the discrimination of the fine-grained features. 
With such feature arrangement module, the model can alleviate the intra-class variance.
Different from \cite{huang2019low}, in this work, we explicitly make full use of the spatial dependencies between the support and query pairs. We propose to generate the attention map based on the pair-wise similarities and reformulate the support image spatial features without external supervision. The proposed method achieves superior performances over \cite{huang2019low}. 
Moreover, to address the low inter-class variance challenge in FG images, existing models~\cite{wei2019piecewise,zhang2019power,li2019CovaMNet,wertheimer2019few,huang2019compare,huang2019low} usually adopt second-order feature extraction. Different from prior works, we propose to integrate the local compositional concept representations into global pair-wise bilinear pooling operation.

Besides the bilinear-based works, generative models \cite{pahde2018discriminative,tsutsui2019meta,he2018only} are also used to synthesize more samples for the support classes. MAML-based model~\cite{zhumulti} adopts a meta-learning strategy to learn good initial FGFS learners. \textcolor{black}{In \cite{CVPRWorkshop}, the authors revised the hyper-spherical prototype network~\cite{CVPRWorkshop2} by maximally separating the classes while incorporating domain knowledge as informative prior. Thanks to the prior knowledge, \cite{CVPRWorkshop} achieves good performance in FGFS classification.}
In~\cite{w6}, the authors proposed a Spatial Attentive Comparison Network (SCAN) to fuse the support-query features based on selective comparison.

\subsection{Generic Few-Shot Learning}
As we study the FG problem under the few-shot setting, generic few-shot learning is also related to our work.
Previous works of the generic Few-Shot (FS) learning are conducted from various perspectives, such as learning with memory~\cite{munkhdalai2017meta,santoro2016meta}, which leverages recurrent neural networks to store the historical information; learning from fine-tuning~\cite{chen2019closer,pmlr-v70-finn17a,rajeswaran2019meta,Sachin2017}, which designs a meta-learning framework to obtain well initial weights for the neural network; learning to compare~\cite{li2019DN4,li2019CovaMNet,snell2017prototypical,Sung_2018_CVPR,vinyals2016matching,zhang2019few}, \textit{etc.} Among which, learning to compare is the most widely used~\cite{gidaris2018dynamic,Hao_2019_ICCV,li2019DN4,li2019CovaMNet,snell2017prototypical,Sung_2018_CVPR,vinyals2016matching,wertheimer2019few,wu2019parn,zhang2019power,CSVT2020}.
In general, learning to compare methods can be divided into two modules: the feature embedding and the similarity measurement. 
By adopting the episode training mechanism~\cite{vinyals2016matching},
these approaches optimize the transferable embedding of both auxiliary data and target data. Then, the query images can be identified by the distance-based classifiers~\cite{Hao_2019_ICCV,li2019DN4,liu2019fewTPN,snell2017prototypical,Sung_2018_CVPR,vinyals2016matching}. 
Most recently, \cite{Hao_2019_ICCV,li2019DN4,wu2019parn} focused on exploring regional information for an accurate similarity comparison.
However, these generic FS methods are not designed to address the high intra-class yet low inter-class variance issue in the FGFS problem. 
In this paper, we aim at tackling the FG classification from the class variance perspective, \textit{i.e.,} we propose to capture a more robust representation of images by simultaneously eliminating the intra-class variations through feature alignment and enhancing the inter-class discrimination by adopting the group pair-wise second-order feature extraction. Thus the proposed method outperforms the generic few-shot models. \textcolor{black}{Based on our analysis, the TOAN framework can also be extended to other weakly-supervised tasks, such as objection detection \cite{w3,w4}, localization \cite{w1,w2}, and segmentation \cite{w5}, \textit{etc.}, where only the image-level supervision is available. The intra-class and inter-class variances in these tasks can be modeled by the proposed target-oriented matching mechanism and global pair-wise bilinear pooling operation, respectively. We will conduct further investigations in our future works.}
\section{Method}\label{Method}

\subsection{Problem Definition} \label{def}
In the FGFS task, we have a small labeled support set $S$ of $C$ different classes. Given an unlabeled query sample $x_q$ from the query set $Q$, the goal is to assign the query to one of the $C$ support classes. This target dataset $\mathcal{D}$ is defined as:
\begin{equation}
\begin{split}
&\mathcal{D} = \left\{  { S } = \left\{ \left({x} _ { s } ,  {y} _ { s } \right) \right\} _ { s = 1 } ^ { K \times C } \right\} \cup \left\{ { Q } = \left\{  {x} _ { q } \right\} _ { q = 1 } ^ { P } \right\}, \\
& {y} _ { s } \in \{ 1, 2, \cdot\cdot\cdot~, C \} ,   { x } \in   \mathbb { R } ^ { N }, P \gg K \times C,
\end{split}
\end{equation} 
where {$x_s$ and $y_s$} denote the feature and label of a support image, respectively. 
The support set and query set share the same label space. If $S$ contains $K$ labeled samples for each of $C$ categories, the task is noted as a $C$-way-$K$-shot problem.

It is far from obtaining an ideal classifier with the limited annotated $S$. Therefore, FGFS models usually utilize a fully annotated dataset, which has the similar data distribution but disjoint label space with $\mathcal{D}$ as an auxiliary dataset $\mathcal{A}$. 
To make full use of the auxiliary set, we follow the widely used episode training strategy~\cite{vinyals2016matching} as our meta-training framework. Specifically, at each training iteration, one support set $\mathcal{A_{\mathcal{S}}}$ and one query set $\mathcal{A_{\mathcal{Q}}}$ are randomly sampled from the set $\mathcal{A}$ to construct a meta-task, where $\mathcal{A_{\mathcal{S}}}$ contains $K \times C$ samples from $C$ different classes. In this way, each training task can mimic the target few-shot problem with the same setting. By adopting thousands of these meta-training operations, the model can transfer the knowledge from the auxiliary $\mathcal{A}$ to the target dataset $\mathcal{D}$.
\begin{figure*}[t]
\begin{center}
\includegraphics[width=0.85\linewidth]{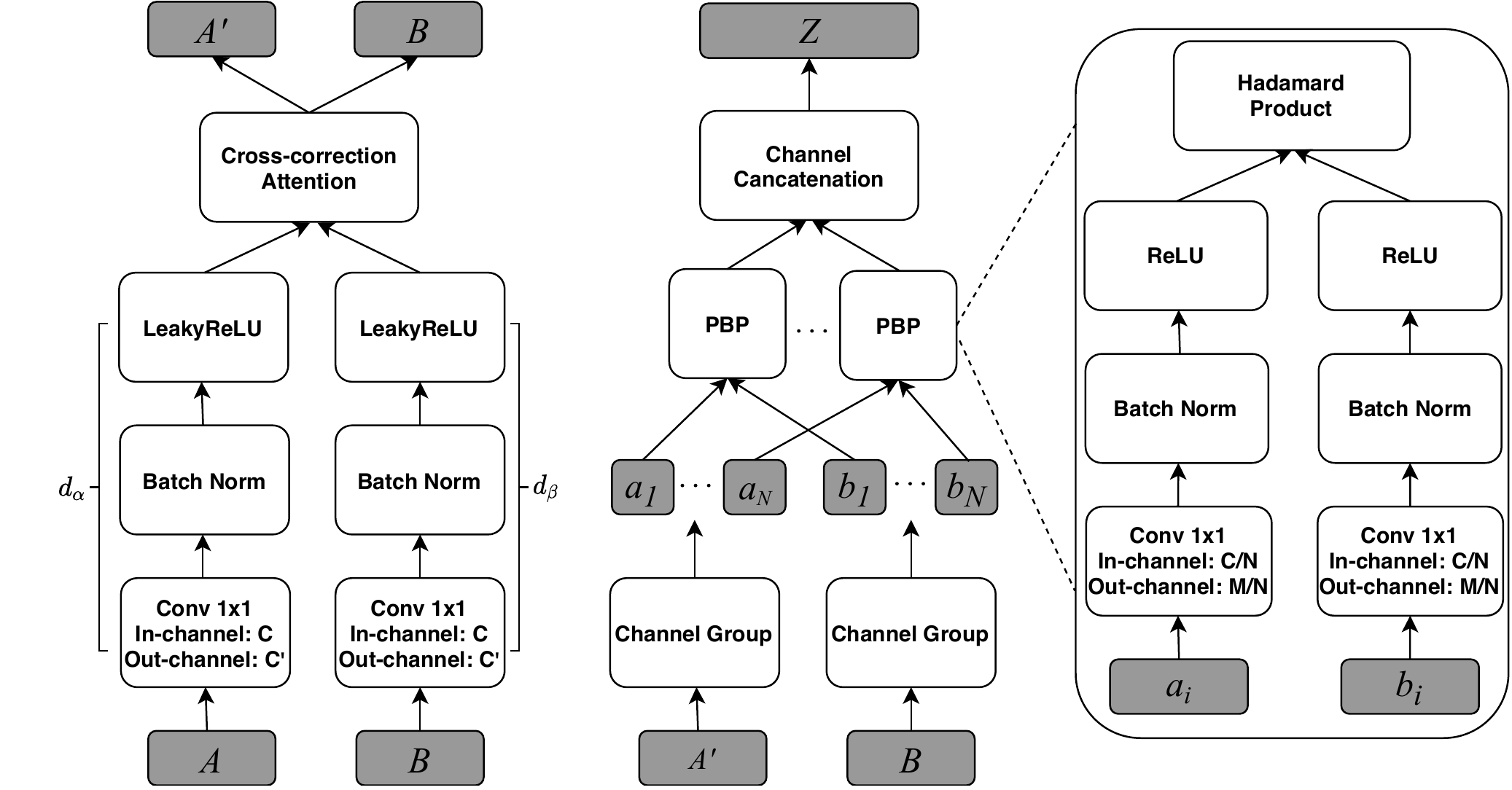}
\end{center}
   \caption{The architecture of fine-grained relation extractor, the left figure denotes TOMM, and the right one represents the GPBP operation. $A, B$ indicate the embedded support sample and query sample, $Z$ is the fine-grained relation.}
\label{architecture}
\end{figure*}

\subsection{The Proposed TOAN}
Given a support image set $S = \left\{ \left({x} _ { s } ,  {y} _ { s } \right) \right\} _ { s = 1 } ^ { K \times C } = \left\{ \{ x^{(1)}_{1}, \cdot\cdot\cdot~,x^{(1)}_{K} \},\cdots,\{ x^{(C)}_{1}, \cdot\cdot\cdot~,x^{(C)}_{K} \}   \right\}$\footnote{For a clear understanding, we group the support set $S$ into $C$ subsets.}, where $x^{(t)}_{K}$ is the $K$-th sample in class $t$, and a query image set $Q = \left\{ {x} _ {q} \right\} _ { q = 1 } ^ { P } $, the learning to compare FS model generally consists of two parts: feature embedding module $f_\theta$ and the comparator $C_\phi$, which can be described as:
\begin{equation}
\text{FS}(S,x_q) =  C_{\phi} \circ f_\theta(S,x_q),
\label{FS_model}
\end{equation}
where $\circ$ denotes the operator of the function composition, $f_\theta$ aims to learn the feature embedding of raw images, and $C_\phi$ is the classifier. However, this framework cannot capture the subtle difference in FG data. Accordingly, FGFS models~\cite{li2019CovaMNet,wei2019piecewise,wertheimer2019few,zhang2019power} incorporate a high-order feature generation module to address the low inter-class variance. However, the sizeable intra-class variance issue is less considered in these methods.

To this end, we propose the Target-Oriented Alignment Network (TOAN) to jointly tackle these issues through a deep fine-grained relation extractor $g_\omega$.
Fig. \ref{framework} illustrates the workflow of the proposed model: 
\begin{equation}
\text{TOAN}(S,x_q) =  C_{\phi} \circ g_\omega \circ f_\theta(S,x_q) = C_{\phi}(Z_{S,q}),
\label{TOAN_def}
\end{equation}
where the comparator $C_\phi$ assigns each $x_q$ to its nearest category in $S$ according to the fine-grained relation $Z_{S,q}$, which is generated by applying $g_\omega$ on the embedded features $f_\theta(S)$ and $f_\theta(x_q)$ as:
\begin{equation}
\begin{split}
& ~Z_{S,q} = g_\omega(f_\theta(S,x_q)) \\
& \quad~~~~ = \text{GPBP} \circ \text{TOMM}(f_\theta(S,x_q))\\
& \quad~~~~ = \text{GPBP}(\{A_1, \cdot\cdot\cdot~,A_C \},B),
\label{Z_def}
\end{split} 
\end{equation}
where $g_\omega$ is composed of two parts, TOMM and GPBP respectively. 
TOMM is designed to generate the query image feature $B$ and a set of support class prototypes $\{A_1, \cdot\cdot\cdot~,A_t, \cdot\cdot\cdot~,A_C\}$, (\textit{e.g.,} $A_t$ represents the prototype of class $t$), which are spatially matched in the embedding space.
GPBP focuses on extracting the second-order features from the aligned support class prototypes and query features. 

\subsubsection{Fine-grained Relation Extraction}
\paragraph{TOMM (Target-Oriented Matching Mechanism)}
\textcolor{black}{As is discussed in the introduction, it is hard to select a support sample as the template to fully represent its category. Moreover, traditional template-based methods~\cite{tmp1,tmp2,tmp3,tmp4} often require a large amount of labeled data. To address the intra-class variance issue with limited training data, we choose each query sample (target) as the template to orient all support samples. 
Since different sub-classes belong to the same entry-level class, samples from those classes often share similar appearances. The similarities of same parts among sub-classes are higher than those in different parts from the same class. Therefore, we instantiate TOMM using the cross-correction attention.}
That is, for a pair of support image $x^{(t)}_s$ and query image $ x_q $, TOMM is expressed as:

\begin{equation}
\begin{split}
& A^{(t)}_s, B = f_{\theta}(x^{(t)}_s, x_q) \in \mathbb { R } ^ { c \times hw }, \\
& d_{\alpha}: A^{(t)}_s  \longrightarrow \mathbb{R} ^ {c' \times hw}, d_{\beta}: B  \longrightarrow \mathbb{R} ^ {c' \times hw},\\
\label{TOMM-1}
\end{split} 
\end{equation}
where $c$ indicates the channel number of the convolutional feature map and $h,w$ denote the size of the feature map. $A^{(t)}_s$ and $B$ are the embedded support and query features. \textcolor{black}{$d_{\alpha}$ and $d_{\beta}$ are two convolutional sub-networks that capture the task-agnostic similarity between two features ($c' \leq c$).} The aligned support feature $A'^{(t)}_s$ and the support class prototype $A_t$ is computed as follows:
\begin{equation}
\begin{split}
& (A'^{(t)}_s)^{T} = \text{Softmax}(\frac{d_{\beta}(B)^{T}d_{\alpha}(A^{(t)}_s)}{\sqrt{c'}})(A^{(t)}_s)^{T},\\
& A_t = \frac{1}{K} \sum_{s=1}^{K} A'^{(t)}_s,
\label{TOMM-2}
\end{split} 
\end{equation}
where the $\text{Softmax}(\cdot)$ operates in a row-wise way. $A^{(t)}_s$ is transformed to $A'^{(t)}_s$, where the similarity of each spatial position between $A'^{(t)}_s$ and $B$ reaches maximal. 
By averaging all aligned features in the given support class, TOMM obtains spatially matched support-class prototypes and query features. Consequently, the intra-class variance in each class is reduced.
{As is shown in Fig. \ref{framework}, the red support bird's embedded features are reformulated according to the query support bird through our TOMM module. It explicitly transforms the `posture' of support image to match the query ones.
}

It is worth noting that, for generic FS tasks, since the inter-class variance is relatively large, the cross-correction attention is used to locate the closest features to classify different classes \cite{hou2019cross,wu2019parn,Hao_2019_ICCV}. However, in FG classification, the inter-class variance is relatively subtle, yet much higher intra-class variance exists. Those closest features between query-gallery pairs often perform poorly compared with generic FS. 
Therefore, we propose to use the cross-correction mechanism to align the feature pairs instead of finding the closest features. Specifically, we explicitly transfer the support image features to match the query ones spatially. 

\paragraph{GPBP (Group Pair-wise Bilinear Pooling)}
Semantic compositional information plays an important role in FG tasks, as the discriminative information always exists in some small parts. However, current FGFS models~\cite{wei2019piecewise,zhang2019power,wertheimer2019few,huang2019low} focus on learning the FG features from the global view. Moreover, studies show that high-level convolutional channels represent specific semantic patterns~\cite{zhang2017interleaved,zheng2019learning,zheng2017learning}.
To this end, we propose to combine compositional concept representations into the second-order feature extraction to generate more discriminative features for FGFS.

GPBP is composed of the convolutional channel grouping operation followed by the pairwise bilinear feature extraction. Given a pair of support class feature $A_t \in \mathbb{R}^{c \times hw}$ and query image feature $B \in \mathbb{R}^{c \times hw}$, we define the semantic grouping operation as follows:
\begin{equation}
\begin{split}
&\hat{A} = \text{Group}(A_t), \hat{B} = \text{Group}(B), \\
& \text{Group}(\cdot):I \longrightarrow [i_1; \cdot\cdot\cdot~; i_k; \cdot\cdot\cdot~; i_N], \\
&I \in \mathbb{R}^{c \times hw}, i_k \in \mathbb{R}^{\frac{c}{N} \times hw},
\label{group}
\end{split} 
\end{equation}
where $\text{Group}(\cdot)$ converts the original feature into $N$ different groups along the channel dimension, each of these feature groups contains $\frac{c}{N}$ channels, which corresponds to a semantic subspace~\cite{hu2019weakly}. For $\hat{A} = [a_1; \cdot\cdot\cdot~; a_k; \cdot\cdot\cdot~; a_N]$ and $\hat{B} = [b_1; \cdot\cdot\cdot~; b_k; \cdot\cdot\cdot~; b_N]$, we define a bilinear feature $z_p$ of $a_k$ and $b_k$ as:
\begin{equation}
\begin{split}
& z_p = \text{Bilinear}(a_k,b_k,W_{kp}) \in \mathbb{R}^{ 1 \times hw} \\
& \quad = [(a^1_k)^T W_{kp}b^1_k, \cdot\cdot\cdot~,(a^{hw}_k)^T  W_{kp}b^{hw}_k],
\label{pbp-1}
\end{split} 
\end{equation}
where $a^i_k, b^i_k \in \mathbb{R}^{\frac{c}{N}\times 1}$ represent the spatial features of $a_k$ and  $b_k$ in the given position $i$. $W_{kp} \in \mathbb{R}^{\frac{c}{N} \times \frac{c}{N}}$ is a projection matrix that fuses $a^i_k$ and $b^i_k$ into a scalar. By adopting $W_{kp}$ on each spatial position of feature pairs, a bilinear feature $z_p \in \mathbb{R}^{1 \times hw}$ is obtained. For each channel group $k$, GPBP learns $\frac{M}{N}$ projection matrices ($M$ is the dimension of the final bilinear feature), and then we concatenate these scalars to generate a fine-grained relation:
\begin{equation}
\begin{split}
& Z_k = [z_1; \cdot\cdot\cdot~;z_p; \cdot\cdot\cdot~;z_{\frac{M}{N}}] \in \mathbb{R}^{\frac{M}{N} \times hw}.
\label{pbp-2}
\end{split} 
\end{equation}
After obtaining the fine-grained relations of each group, we combine them into the final relation $Z$ \footnote{For brevity, we omit the subscript of $Z_{\hat{A},\hat{B}}$} as:
\begin{equation}
\begin{split}
& Z = [Z_1; \cdot\cdot\cdot~;Z_k; \cdot\cdot\cdot~;Z_N] \in \mathbb{R}^{M \times hw},
\label{pbp-3}
\end{split} 
\end{equation}
where $M$ is the final dimension of $Z$. Similar to \cite{kim2016hadamard,Yu_2018_ECCV}, we adopt a low-rank approximation of $W_{kp}$ to reduce the number of parameters for regularization:
\begin{equation}
\begin{split}
& z_p = \text{Bilinear}(a_k,b_k,W_{kp}) \\
& \quad = [(a^1_k)^T W_{kp}b^1_k, \cdot\cdot\cdot~,(a^{hw}_k)^T  W_{kp}b^{hw}_k]\\
& \quad = [(a^1_k)^T U_{kp}V_{kp}^T b^1_k, \cdot\cdot\cdot~,(a^{hw}_k)^T U_{kp}V_{kp}^T b^{hw}_k]\\
& \quad = [U^T_{kp}a^1_k \odot V^T_{kp}b^1_k, \cdot\cdot\cdot~, U^T_{kp}a^{hw}_k \odot V^T_{kp}b^{hw}_k ] \\
& \quad = (U^T_{kp}[a^1_k, \cdot\cdot\cdot~,a^{hw}_k]) \odot (V_{kp}^T [b^1_k, \cdot\cdot\cdot~,b^{hw}_k])\\
& \quad = (U^T_{kp}a_k) \odot (V_{kp}^T b_k),
\label{lowrank}
\end{split} 
\end{equation}
where $U_{kp} \in \mathbb{R}^{\frac{c}{N} \times 1}$, $V_{kp} \in \mathbb{R}^{\frac{c}{N} \times 1}$, and $\odot$ denotes the Hadamard product. 

\subsubsection{Comparator}
After capturing the comparative bilinear feature of query image $i$ and support class $j$, 
the comparator is defined as:
\begin{equation}
\begin{split}
    & C_\phi(\cdot):{Z}_{i,j} \in \mathbb{R}^{M \times hw} \longrightarrow \mathbb{R}^{1},  \\
    & j \in \{ 1, 2, \cdot\cdot\cdot~, C \}, i  \in \{ 1, 2, \cdot\cdot\cdot~, P \}, \\
\end{split}
\label{comparator}
\end{equation}
where $C_\phi$ learns the distance between the support class $j$ and query image $i$, that is, for each query $i$, the comparator generates similarities from $C$ support categories. The query image is assigned to the nearest category.
Same as \cite{Sung_2018_CVPR,huang2019compare}, we use the MSE loss as our training loss to regress the predicted label to the ground-truth. 

\subsection{Network Architecture}
\noindent\textbf{Feature Embedding Module:} In FGFS and FS tasks, $f_{\theta}$ can be any proper convolutional neural network such as ConvNet-64~\cite{chen2019closer,Sung_2018_CVPR,vinyals2016matching}. 
\noindent\textbf{Fine-grained Relation Extractor:} We show the architecture details of the fine-grained relation extraction module in Fig. \ref{architecture}.
\textit{TOMM}: To construct $d_{\alpha}$ and $d_{\beta}$, we use a convolutional layer with a $1 \times 1$ kernel followed by the batch normalization and a LeakyReLU layer. The Cross-correction Attention is implemented using Eq.~(\ref{TOMM-2}).
\textit{GPBP}: For the channel grouping, we split the embedded feature map into $N$ groups along the channel dimension. Pairwise bilinear pooling (PBP) consists of a convolutional layer with a $1\times1$ kernel followed by the batch normalization and a ReLU layer. Then the Hadamard product operation is applied to generate the final bilinear features.

\noindent\textbf{Comparator}:
The comparator consists of two convolutional blocks and two fully-connected layers. Each block contains a $3 \times 3$ convolution, a batch normalization, and a ReLU layer. The activation function of the first fully connected layer is ReLU, where the Sigmoid function is added after the last fully connected layer to generate similarities of input pairs. 

\section{Experiment}\label{exp}
\begin{table}[t]
	\begin{center}
		\fontsize{10}{13.5}\selectfont
		\caption{The dataset splits. $C_{all}$ is the total number of categories. $C_{\mathcal{ T }}$ is the number of categories in the target datasets. $C_{\mathcal{ A}\_{tr}}$ and $C_{\mathcal{ A}\_va}$ note the training class number and validation category number in the auxiliary datasets separately.} \label{split}
		\begin{tabular}{ | c | c | c | c | c| } \hline
			\text {Dataset} & { \text{CUB}  } & { \text { DOGS } } & { \text { CARS } } & {\text{NABirds}} \\ 
			\hline
			$C _ {all}$ & { 200 } & { 120 } & { 196 } &{555} \\ 
			$C _ { \mathcal { T } }$ & { 50 } & { 30 } & { 49 }&{139} \\ 
			$C _ { \mathcal{ A}\_{tr}  }$ & { 120 } & { 70 } & { 130 }&{350} \\ 
			$C _ { \mathcal{ A}\_{va}  }$ & { 30 } & { 20 } & { 17 }&{66} \\ 
			\hline 
		\end{tabular} 
	\end{center}
\end{table}

\subsection{Dataset}
We evaluated the proposed method on four datasets: (1) Caltech-UCSD Birds-200-2011 (CUB)~\cite{WahCUB_200_2011}, which contains 11,788 images of 200 sub-categories of birds; (2) Stanford Dogs (DOGS)~\cite{KhoslaYaoJayadevaprakashFeiFei_FGVC2011}, which consists of 20,580 images from 120 dog species; (3) Stanford Cars (CARS)~\cite{KrauseStarkDengFei-Fei_3DRR2013}, which has 196 categories of cars and a total number of 16,185 images; and (4) North America Birds (NABirds)~\cite{Horn_2015_CVPR}, which consists of 48,562 bird images from 555 bird species.
For fair comparisons, we follow the latest data splits~\cite{huang2019low,li2019DN4,li2019CovaMNet} of FG benchmarks in DN4~\cite{li2019DN4}, as Table~\ref{split} shows.
Since our work focuses on investigating the fine-grained image categorization with limited labeled data rather than generic few-shot learning, we conducted experiments on widely used fine-grained image benchmarks rather than few-shot datasets, such as MiniImageNet~\cite{vinyals2016matching}.

\begin{table}[t]  
	\centering  
	\fontsize{9.8}{14.5}\selectfont  
	\caption{Re-implementations of meta-learning frameworks for the validation purpose.} \label{Val}
	\begin{tabular}{|c|c|c|}  
		\hline
		\multirow{2}{*}{Methods} & 	\multicolumn{2}{c|}{MiniImageNet (\%)} \cr\cline{2-3}
		&1-shot & 5-shot  \\
		\hline
		RelationNet~\cite{Sung_2018_CVPR}  & 50.44$\pm$0.82 & 65.32$\pm$0.70 \\ 
		RelationNet, ours  & 51.87$\pm$0.45  & 64.75$\pm$0.57 \\ \hline
		\hline
		ProtoNet~\cite{snell2017prototypical}  & 49.42$\pm$0.78 & 68.20$\pm$0.66 \\ 
		ProtoNet, ours  & 47.57$\pm$0.63  & 66.21$\pm$0.58 \\ \hline
		\hline
		MatchingNet~\cite{vinyals2016matching}  & 43.56$\pm$0.84 & 55.31$\pm$0.73 \\ 
		MatchingNet, ours  & 48.90$\pm$0.62  & 65.67$\pm$0.55 \\
	   \hline
	\end{tabular}  
\end{table}

\begin{table*}[t]
	\centering  
	\fontsize{7}{15}\selectfont  
	\caption{Fine-grained Few-shot classification accuracy (\%) comparisons on four FG benchmarks. All results are with $95\%$ confidence intervals where reported. We highlight the best and second-best methods.} \label{ALL}
	\begin{tabular}{|c|c|c|c|c|c|c|c|c|c|c|}  
		\hline
		\multirow{2}{*}{Methods} & \multirow{2}{*}{Type} & \multirow{2}{*}{Backbone} &	\multicolumn{2}{c|}{CUB}&\multicolumn{2}{c|}{CARS }&\multicolumn{2}{c|}{DOGS}&\multicolumn{2}{c|}{NABirds}  \cr\cline{4-11}
		&&&1-shot & 5-shot & 1-shot & 5-shot&1-shot & 5-shot & 1-shot & 5-shot \\
		\hline  \hline
		PCM~\cite{wei2019piecewise} & FGFS & AlexNet & 42.10$\pm$1.96 & 62.48$\pm$1.21& 29.63$\pm$2.38 & 52.28$\pm$1.46 & 28.78$\pm$2.33 & 46.92$\pm$2.00  & - & -\\ \hline 
		PABN+$_{cpt}$~\cite{huang2019low} & FGFS & ConvNet-64 & 63.36$\pm$0.80 & 74.71$\pm$0.60 & 54.44$\pm$0.71 & 67.36$\pm$0.61 & 45.65$\pm$0.71 & 61.24$\pm$0.62 & 66.94$\pm$0.82 & 79.66$\pm$0.62  \\ \hline 
		LRPABN$_{cpt}$~\cite{huang2019low} & FGFS & ConvNet-64 &  {{{63.63$\pm$0.77}}} & {{76.06$\pm$0.58}} &  {{{60.28$\pm$0.76}}} & {{73.29$\pm$0.58}} & 45.72$\pm$0.75 & 60.94$\pm$0.66 &  {{{67.73$\pm$0.81}}} & {{81.62$\pm$0.58}}  \\ \hline 
		SoSN~\cite{zhang2019power} & FGFS & ConvNet-64 & 63.95$\pm$0.72 & 78.79$\pm$0.60 & 62.84$\pm$0.68&  75.75$\pm$0.52  & 48.01$\pm$0.76 & 64.95$\pm$0.64 & 69.53$\pm$0.77 & 83.87$\pm$0.51 \\ \hline 
		\hline
		MatchingNet~\cite{vinyals2016matching} & FS &  ConvNet-64~ & 57.59$\pm$0.74 & 70.57$\pm$0.62 & 48.03$\pm$0.60  & 64.22$\pm$0.59 & 45.05$\pm$0.66  & 60.60$\pm$0.62   & 60.70$\pm$0.78  & 76.23$\pm$0.62 \\ \hline 
		ProtoNet~\cite{snell2017prototypical} & FS & ConvNet-256 & 53.88$\pm$0.72  & 70.85$\pm$0.63 & 45.27$\pm$0.61 & 64.24$\pm$0.61 & 42.58$\pm$0.63 & 59.49$\pm$0.65 & 55.85$\pm$0.78 & 75.34$\pm$0.63 \\ \hline
		RelationNet~\cite{Sung_2018_CVPR} & FS & ConvNet-64 & 59.82$\pm$0.77  & 71.83$\pm$0.61 & 56.02$\pm$0.74 & 66.93$\pm$0.63 & 44.75$\pm$0.70 & 58.36$\pm$0.66 & 64.34$\pm$0.81 & 77.52$\pm$0.60 \\ \hline
		CovaMNet~\cite{li2019CovaMNet} & FS & ConvNet-64 & 58.51$\pm$0.94 &71.15$\pm$0.80 &56.65$\pm$0.86 &71.33$\pm$0.62 & {{{49.10$\pm$0.76}}}& {{63.04$\pm$0.65}}&60.03$\pm$0.98 & 75.63$\pm$0.79  \\ \hline
		DN4~\cite{li2019DN4} & FS & ConvNet-64 & 55.60$\pm$0.89 & {{{77.64$\pm$0.68}}} &{{59.84$\pm$0.80}} & {{\textbf{88.65$\pm$0.44}}} &45.41$\pm$0.76 &  {{{63.51$\pm$0.62}}} & 51.81$\pm$0.91 &  {{{83.38$\pm$0.60}}}  \\ \hline
		\hline
		TOAN & FGFS & ConvNet-64 &\textbf{65.34$\pm$0.75}  & \textbf{80.43$\pm$0.60} & \textbf{65.90$\pm$0.72} & {84.24$\pm$0.48} & \textbf{49.30$\pm$0.77} & \textbf{67.16$\pm$0.49}  & \textbf{70.02$\pm$0.80} & \textbf{85.52$\pm$0.50}\\ \hline
		TOAN:ResNet & FGFS & ResNet-256 & \textbf{67.17$\pm$0.81}  & \textbf{82.09$\pm$0.56} & \textbf{76.62$\pm$0.70} & \textbf{89.57$\pm$0.40} & \textbf{51.83$\pm$0.80} & \textbf{69.83$\pm$0.66}  & \textbf{76.14$\pm$0.75} & \textbf{90.21$\pm$0.40}\\
		\hline 
	\end{tabular} 
\end{table*}

\subsection{Experimental Setting}
All experiments were conducted in the 5-way-1-shot and 5-way-5-shot fashions. During each episode of training and testing, we randomly selected five categories to construct a FGFS task. For the 5-way-1-shot setting, we randomly sampled 5$\times$1$+$5$\times$15 = 80 images from the selected categories, where there are one support image and 15 query images in each class. Similarly, we randomly chose 5$\times$5$+$5$\times$10 = 75 images to set up the 5-way-5-shot experiment.
We resized the input image to 84$\times$84 and trained models from the scratch using the Adam optimizer \cite{kingma2015adam}. The initial learning rate is 0.001. We set the group number of GPBP as four and the bilinear feature dimension as 1024. \textcolor{black}{ Moreover, we fixed the output channel of $d_{\alpha}$ and $d_{\beta}$ as 64 ($c' = 64$ in Eq.~(\ref{TOMM-1})).} 
\textcolor{black}{In the training stage, we randomly constructed 300,000 episodes from the auxiliary datasets to train proposed models by employing the episodic training mechanism~\cite{vinyals2016matching}. We repeated the testing ten times and reported the top one mean accuracy. During each testing iteration, we randomly selected 1000 episodes from target datasets. Moreover, we also report the 95\% confidence intervals of all models, which is the same as recent works~\cite{vinyals2016matching,snell2017prototypical,Sung_2018_CVPR,li2019DN4,Gidaris_2019_ICCV}.}
For fair comparisons, all experiments were conducted without additional data augmentations~\cite{li2019DN4,li2019CovaMNet}, \textcolor{black}{and we chose the ConvNet-64~\cite{Sung_2018_CVPR} as the main backbone since it has been widely used in most recent works, including PABN~\cite{huang2019low}, LRPABN \cite{huang2019low}, SoSN \cite{zhang2019power}, MatchingNet \cite{vinyals2016matching}, RelationNet \cite{Sung_2018_CVPR}, DN4 \cite{li2019DN4}. For other compared methods, we maintain the same backbones as reported in their original works.} We specify these backbones in Table~\ref{ALL}. In our experiment, the compared methods include:
\paragraph{Baselines}
MatchingNet~\cite{vinyals2016matching}, ProtoNet~\cite{snell2017prototypical}, and RelationNet \cite{Sung_2018_CVPR} are three exemplary few-shot learning methods. For fair comparisons, we re-implemented these methods by referring to the source codes with our experimental settings.
We conducted verification experiments on the MiniImageNet dataset~\cite{vinyals2016matching} with the ConvNet-64 backbone~\cite{Sung_2018_CVPR} to validate the correctness of our re-implementations of these three baselines. 
We present the comparison results of our re-implementations against the original models in Table~\ref{Val}. It can be observed that the classification accuracies of our re-implementations posses no more than 2\% fluctuations. 
These minor margins are mostly caused by the differences in the experimental settings, as \cite{chen2019closer} investigated.

\paragraph{FGFS models}
Since we aim at tackling the fine-grained classification under the few-shot setting, we selected the most related FGFS methods as main comparisons, including the first FGFS model PCM~\cite{wei2019piecewise}. The state-of-the-art models SoSN~\cite{zhang2019power}, and FGFS models PABN+ as well as LRPABN$_{cpt}$ from \cite{huang2019low} for comparison. For PCM, PABN+, and LRPABN$_{cpt}$, we quote the reported results.
For other models, the results on the four benchmarks are obtained from their open-sourced models.

\paragraph{Generic FS models}
It is worth noting that generic FS models can still be applied to fine-grained data. By referring to the first FGFS method \cite{wei2019piecewise}, we selected the most representative ones for comparisons. That is, we compare our model against DN4~\cite{li2019DN4} and CovaMNet~\cite{li2019CovaMNet}. Results on CUB and NABirds are obtained from their open-sourced models, while others are quoted from reported results.

\paragraph{TOAN Family}
First of all, we added the proposed matching mechanism TOMM to FS baseline models to investigate its effectiveness for FGFS tasks, noted as FS+TOMM, where FS can be any one of the baselines. Similarly, GPBP is also plugged into the RelationNet, noted as RelationNet+GPBP.
To investigate the grouping function, we replaced the proposed function by a $1 \times 1$ convolutional layer with a group parameter~\cite{zhang2017interleaved}, noted as {TOAN-GP*}.
Moreover, we removed the task-agnostic transformation $d(\cdot)$ in TOMM, noted as {TOAN-$w/o ~{d(\cdot)}$}, where pairwise similarities are computed directly based on the embedded support feature $A_t$ and query feature $B$.
We also replaced the backbone ConvNet-64 with the deeper ResNet-256\cite{li2019DN4} to study the influence of different backbones, noted as {TOAN:ResNet}. 
Finally, we used a larger $224\times224$ input image size with different backbones to study the effects of the input size for TOAN, noted as {TOAN\_224} and {TOAN:ResNet\_224}, respectively.

\begin{table*}[!hbpt]  
	\centering  
	\fontsize{8.5}{15.0}\selectfont  
	\caption{The ablation study on TOMM. 
    We incorporate each framework with TOMM and observe definite improvements (\%).} \label{TOMM}
	\begin{tabular}{|c|c|c|c|c|c|c|c|c|}  
		\hline
		\multirow{2}{*}{Methods} & 	\multicolumn{2}{c|}{CUB}&\multicolumn{2}{c|}{CARS }&\multicolumn{2}{c|}{DOGS}&\multicolumn{2}{c|}{NABirds}  \cr\cline{2-9}
		&1-shot & 5-shot & 1-shot & 5-shot&1-shot & 5-shot & 1-shot & 5-shot \\
		\hline  \hline
		MatchingNet~\cite{vinyals2016matching} & 57.59$\pm$0.74 & 70.57$\pm$0.62 & 48.03$\pm$0.60  & 64.22$\pm$0.59 & 45.05$\pm$0.66  & 60.60$\pm$0.62   & 60.70$\pm$0.78  & 76.23$\pm$0.62 \\ 
		MatchingNet+TOMM  & \textbf{60.87$\pm$0.78} & \textbf{75.12$\pm$0.61} &\textbf{53.79$\pm$0.72}  &\textbf{72.67$\pm$0.55} &\textbf{47.06$\pm$0.74}  &\textbf{63.22$\pm$0.62}   &\textbf{65.83$\pm$0.75}  &\textbf{80.73$\pm$0.57} \\ 
		           & +3.28  & +4.55  & +5.76  & +8.45  & +2.01 & +2.62 & +5.13 & +4.50 \\\hline
		ProtoNet~\cite{snell2017prototypical} & 53.88$\pm$0.72  & 70.85$\pm$0.63 & 45.27$\pm$0.61 & 64.24$\pm$0.61 & 42.58$\pm$0.63 & 59.49$\pm$0.65 & 55.85$\pm$0.78 & 75.34$\pm$0.63 \\
		ProtoNet+TOMM & \textbf{61.60$\pm$0.76}  & \textbf{75.09$\pm$0.61} & \textbf{52.50$\pm$0.69} & \textbf{68.13$\pm$0.58} & \textbf{46.36$\pm$0.73} & \textbf{61.56$\pm$0.65}  & \textbf{64.77$\pm$0.79} & \textbf{80.84$\pm$0.56}\\
		 & +7.72  & +4.24  & +7.23  & +3.89 & +3.78 & +2.07 & +9.92 & +5.50 \\\hline
		RelationNet~\cite{Sung_2018_CVPR} & 59.82$\pm$0.77  & 71.83$\pm$0.61 & 56.02$\pm$0.74 & 66.93$\pm$0.63 & 44.75$\pm$0.70 & 58.36$\pm$0.66 & 64.34$\pm$0.81 & 77.52$\pm$0.60 \\
		RelationNet+TOMM & \textbf{64.84$\pm$0.77}  & \textbf{79.75$\pm$0.54} & \textbf{62.35$\pm$0.77} & \textbf{81.57$\pm$0.51} & \textbf{47.24$\pm$0.78} & \textbf{65.23$\pm$0.66}  & \textbf{69.55$\pm$0.77} & \textbf{85.01$\pm$0.51}\\
		 & +5.02  & +7.92 & +6.33  & +14.64 & +2.49 & +6.87 & +5.21 & +7.49 \\\hline
	\end{tabular}
\end{table*}

\begin{table*}[!hbpt]  
	\centering  
	\fontsize{8.64}{15.0}\selectfont  
	\caption{Ablation study of GPBP. We obtain absolute improvements (\%) in each model after incorporating GPBP. We also show the results of the whole model TOAN.} \label{GPBP}
	\begin{minipage}{18.5cm}
	\begin{tabular}{|c|c|c|c|c|c|c|c|c|}  
		\hline
		\multirow{2}{*}{Methods} & 	\multicolumn{2}{c|}{CUB}&\multicolumn{2}{c|}{CARS }&\multicolumn{2}{c|}{DOGS}&\multicolumn{2}{c|}{NABirds}  \cr\cline{2-9}
		&1-shot & 5-shot & 1-shot & 5-shot&1-shot & 5-shot & 1-shot & 5-shot \\
		\hline  \hline
		RelationNet \cite{Sung_2018_CVPR} & 59.82$\pm$0.77  & 71.83$\pm$0.61 & 56.02$\pm$0.74 & 66.93$\pm$0.63 & 44.75$\pm$0.70 & 58.36$\pm$0.66 & 64.34$\pm$0.81 & 77.52$\pm$0.60 \\ 
		RelationNet+GPBP  & \textbf{60.00$\pm$0.74} & \textbf{74.01$\pm$0.60}  & \textbf{58.35$\pm$0.73}   & \textbf{73.49$\pm$0.59}  & \textbf{46.45$\pm$0.70} & \textbf{61.70$\pm$0.65}   & \textbf{65.43$\pm$0.81}  & \textbf{80.13$\pm$0.58}  \\ 
		& +0.18  & +2.18 & +2.33  & +6.56 & +1.70 & +3.34 & +1.09 & +2.61 \\\hline
		RelationNet \cite{Sung_2018_CVPR} & 59.82$\pm$0.77  & 71.83$\pm$0.61 & 56.02$\pm$0.74 & 66.93$\pm$0.63 & 44.75$\pm$0.70 & 58.36$\pm$0.66 & 64.34$\pm$0.81 & 77.52$\pm$0.60 \\ 
		TOAN \footnote{TOAN consists of RelationNet, TOMM, and GPBP together.} & \textbf{65.34$\pm$0.75}  & \textbf{80.43$\pm$0.60} & \textbf{65.90$\pm$0.72} & \textbf{84.24$\pm$0.48} & \textbf{49.30$\pm$0.77} & \textbf{67.16$\pm$0.49}  & \textbf{70.02$\pm$0.80} & \textbf{85.52$\pm$0.50}\\
		& +5.52 & +8.60 & +9.88 & +17.31 & +4.55 & +8.80 & +5.68 & +8.00 \\
		\hline
	\end{tabular} 
	\end{minipage}
\end{table*}
\begin{table*}[!hbpt]  
	\centering  
	\fontsize{8.5}{15.5}\selectfont  
	\caption{Ablation study of TOAN for other choices. Few-shot classification results (\%) on four FG datasets.} \label{Other}
	\begin{tabular}{|c|c|c|c|c|c|c|c|c|}  
		\hline
		\multirow{2}{*}{Methods} & 	\multicolumn{2}{c|}{CUB}&\multicolumn{2}{c|}{CARS }&\multicolumn{2}{c|}{DOGS}&\multicolumn{2}{c|}{NABirds}  \cr\cline{2-9}
		&1-shot & 5-shot & 1-shot & 5-shot&1-shot & 5-shot & 1-shot & 5-shot \\
		\hline  \hline
		TOAN-GP*  & 65.80$\pm$0.78 & 79.37$\pm$0.61  & 65.88$\pm$0.74 & 82.69$\pm$0.50   & 50.10$\pm$0.79  & 65.90$\pm$0.68 & 69.48$\pm$0.75 & 85.48$\pm$0.53  \\ \hline
		TOAN-w/o ${d(\cdot)}$ & 64.48$\pm$0.76 & 78.82$\pm$0.59  & 60.02$\pm$0.73   & 81.65$\pm$0.49  & 47.27$\pm$0.72 & 63.98$\pm$0.65 & 68.70$\pm$0.79  & 83.70$\pm$0.53 \\ \hline
		TOAN & {65.34$\pm$0.75}  & {80.43$\pm$0.60} & {65.90$\pm$0.72} & {84.24$\pm$0.48} & {49.30$\pm$0.77} & {67.16$\pm$0.49}  & {70.02$\pm$0.80} & {85.52$\pm$0.50}\\
		\hline
		TOAN\_224 & {69.03$\pm$0.79}  & {83.19$\pm$0.56} & { 69.48$\pm$0.74 } & {87.38$\pm$0.45} & {53.67$\pm$0.80} & {69.77$\pm$0.70}  & {75.17$\pm$0.76} & {88.77$\pm$0.46}\\ \hline
		TOAN:ResNet\_224 & \textbf{69.91$\pm$0.82}  & \textbf{84.86$\pm$0.57} & \textbf{77.25$\pm$0.73} & \textbf{ 91.19$\pm$0.40 } & \textbf{55.77$\pm$0.79} & \textbf{72.16$\pm$0.72}  & \textbf{77.32$\pm$0.70} & \textbf{91.39$\pm$0.41}\\ \hline
	\end{tabular} 
\end{table*}

\begin{table}[t]
\centering  
\fontsize{8.64}{15.0}\selectfont
\caption{Ablation study of TOAN for the output channel size of $d_{
\alpha}, d_{\beta}$. Few-shot classification results (\%) on the CUB data set.} \label{channel}
\begin{tabular}{|c|c|c|}
\hline
\multicolumn{1}{|l|}{\multirow{2}{*}{Num\_Channel $d(c')$}} & \multicolumn{2}{c|}{CUB  data set}                \\ \cline{2-3} 
\multicolumn{1}{|l|}{}                                  & 5-way-1-shot            & 5-way-5-shot            \\ \hline
32                                                      & 65.47$\pm$0.77          & 80.51$\pm$0.60          \\ \hline
64                                                      & 67.17$\pm$0.81          & 82.09$\pm$0.56          \\ \hline
128                                                     & 68.73$\pm$0.80          & 83.88$\pm$0.61          \\ \hline
256                                                     & \textbf{69.40$\pm$0.81} & \textbf{84.01$\pm$0.59} \\ \hline
\end{tabular}
\end{table}

\subsection{Experimental Results}
\begin{figure*}[t]
	\centering
	\subfloat[CUB~\cite{WahCUB_200_2011} Visualization] {\includegraphics[width=0.245\textwidth]{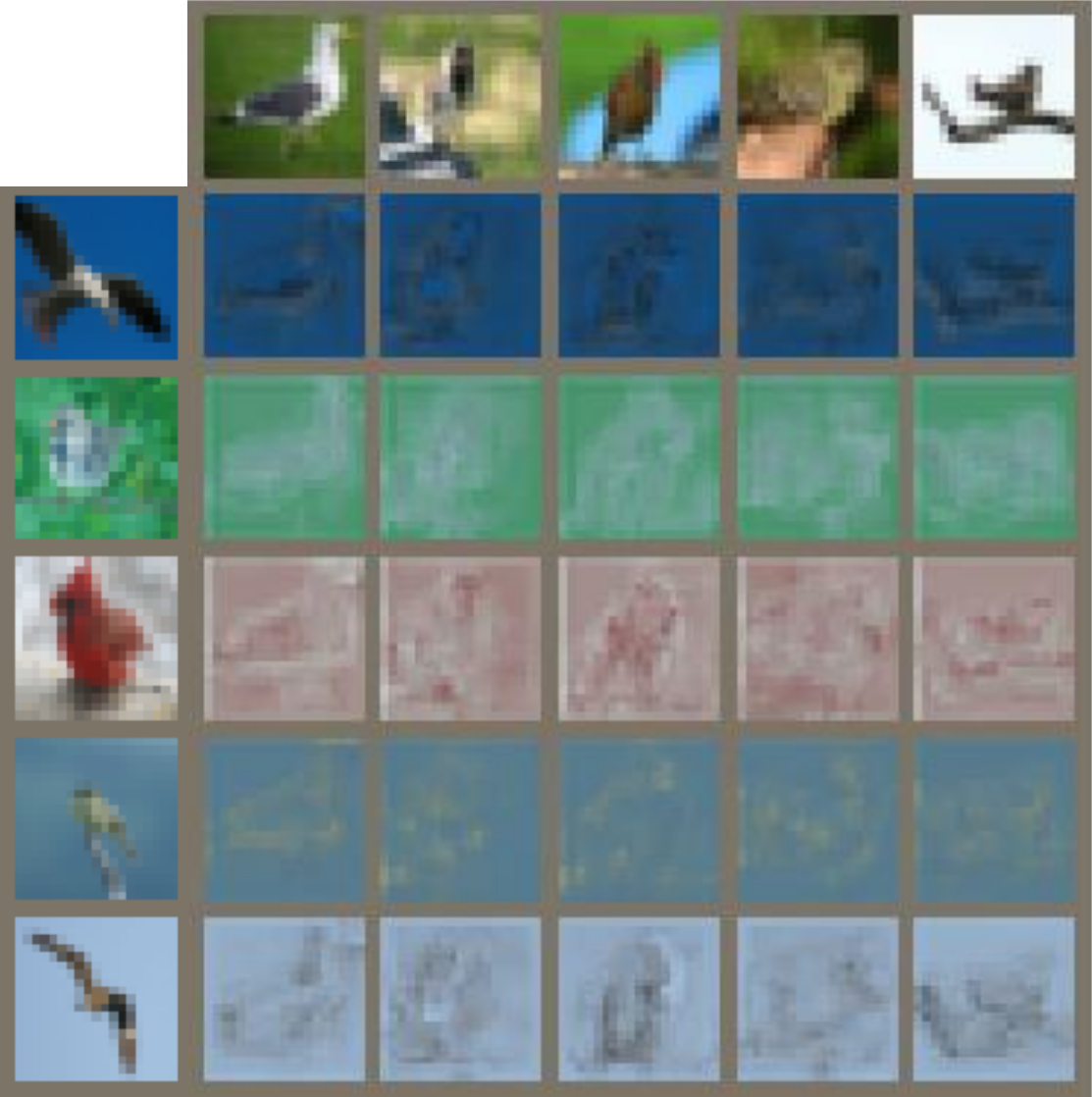} \label{sub001}} 
	\subfloat[CARS~\cite{KrauseStarkDengFei-Fei_3DRR2013} Visualization] {\includegraphics[width=0.245\textwidth]{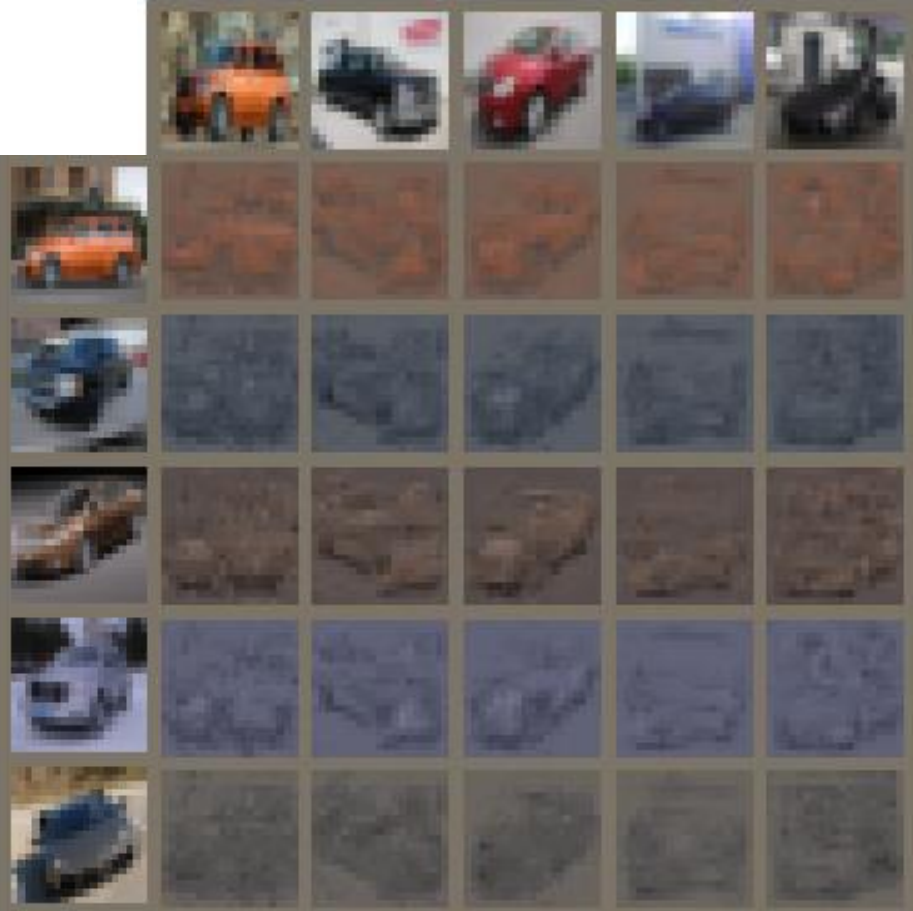} \label{sub002}}
	\subfloat[DOGS~\cite{KhoslaYaoJayadevaprakashFeiFei_FGVC2011} Visualization] {\includegraphics[width=0.245\textwidth]{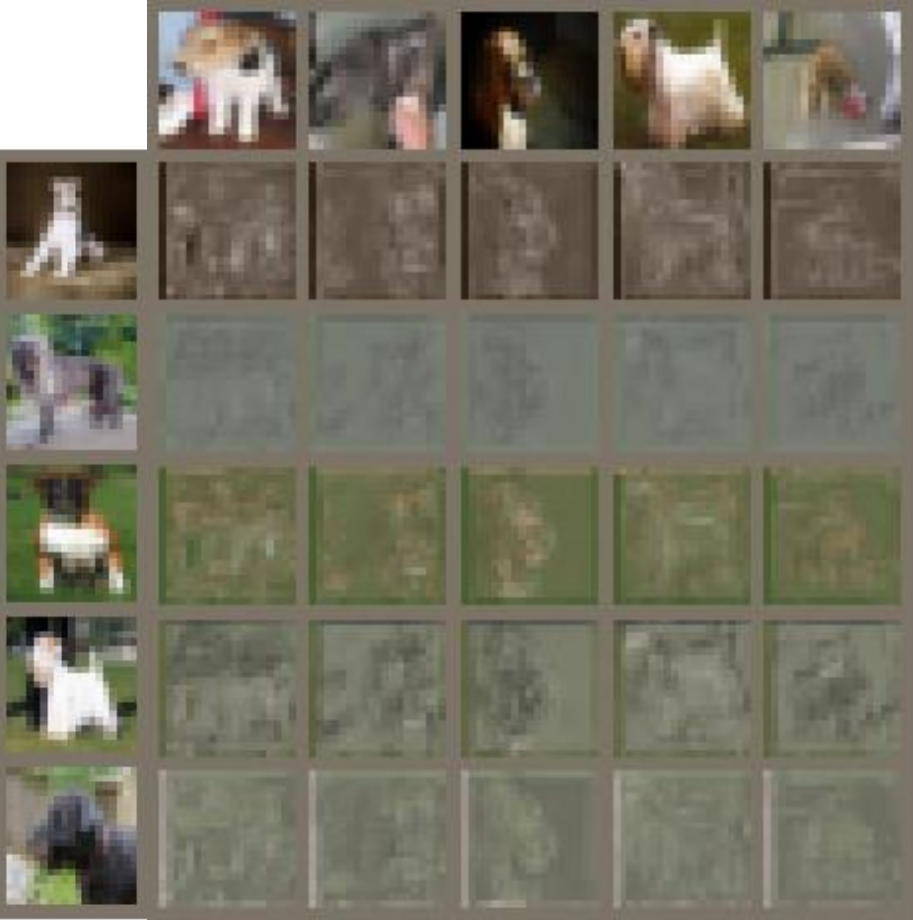} \label{sub003}} 
	\subfloat[NABirds~\cite{Horn_2015_CVPR} Visualization] {\includegraphics[width=0.25\textwidth]{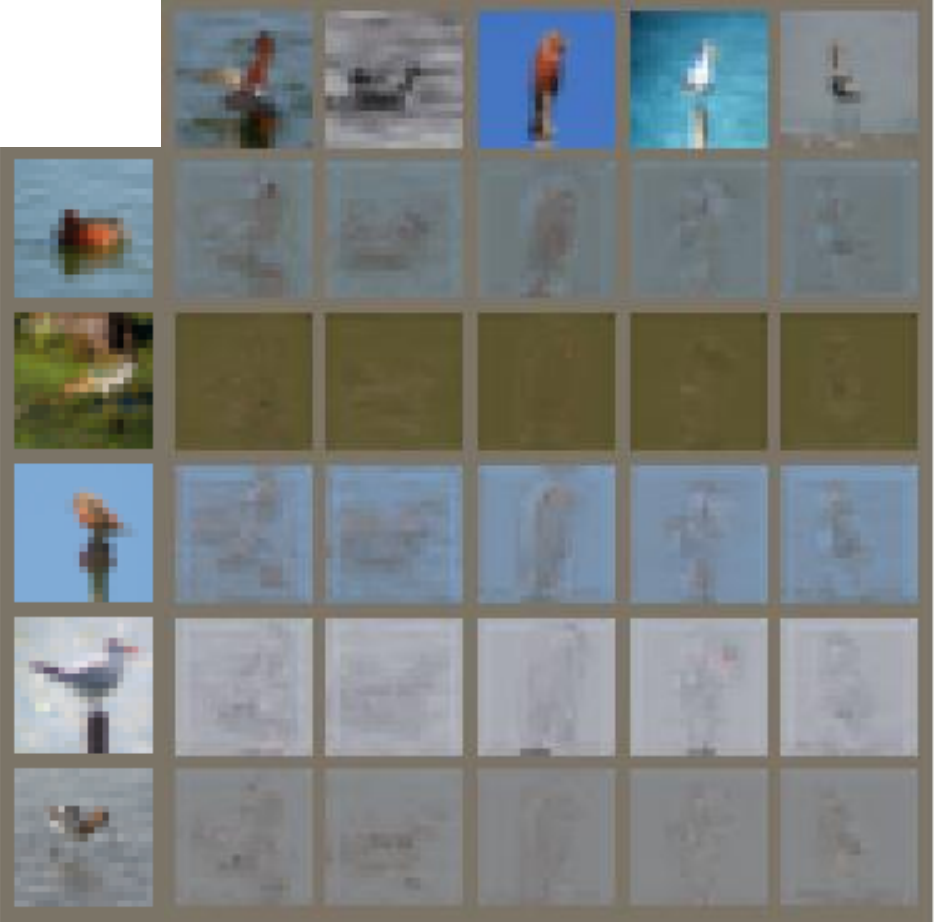} \label{sub004}}
	\caption{TOMM Visualization, the first image in each row (except for the first row) represents the support image, and the remaining images in the row are the aligned results of the support image, which are matched to each query image (in each column from the first row).}
	\label{TOMMSOW}
\end{figure*}

\begin{figure*}[t]
	\centering
	\subfloat[Semantic Grouping Validation.] {\includegraphics[width=0.475\textwidth]{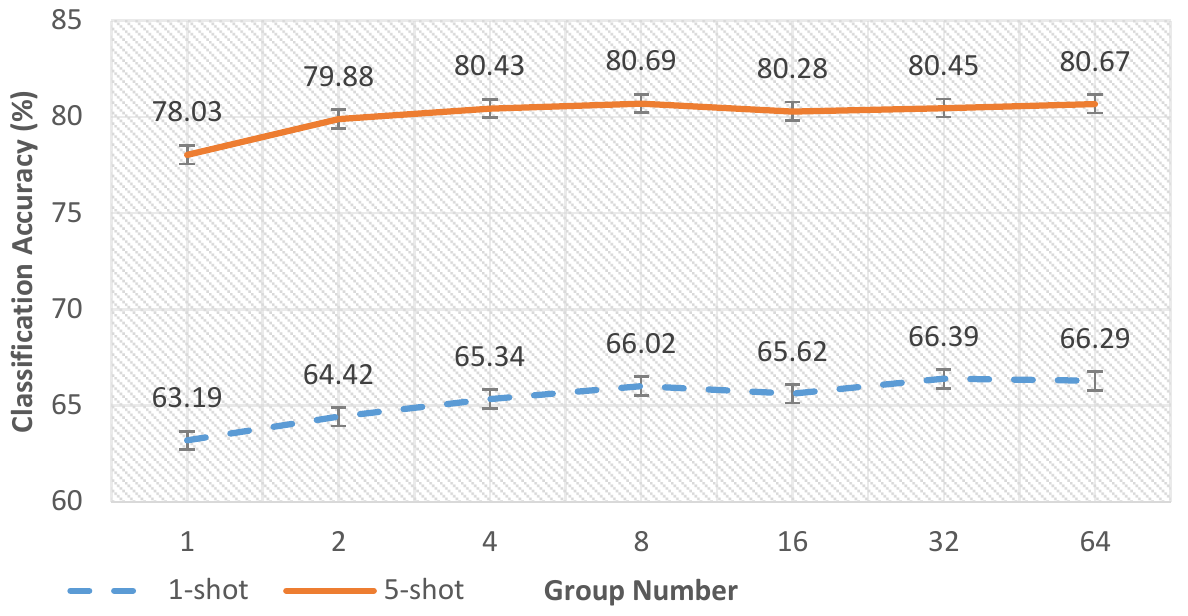} \label{sub2}}
	\subfloat[Features Dimension Selection.] {\includegraphics[width=0.475\textwidth]{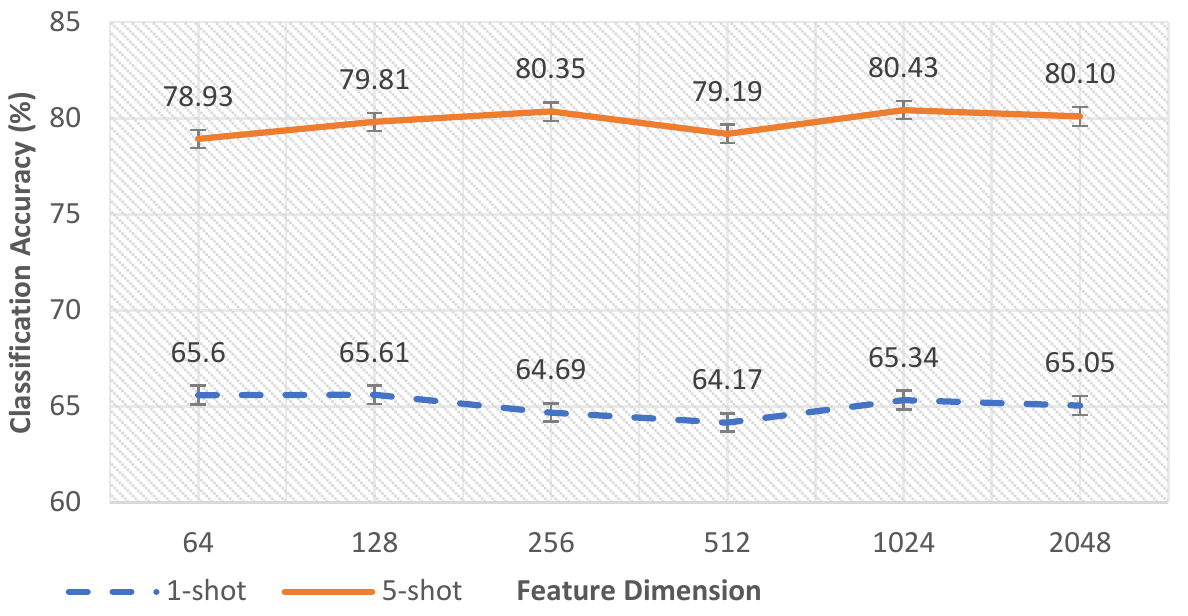} \label{sub1}} 
	\caption{Ablation studies about the proposed GPBP, including semantic channel grouping validation \ref{sub2} and feature dimension selection \ref{sub1}. For each group of validation experiments, we show the 1-shot and 5-shot results.}
	\label{Parameter}
\end{figure*}

\begin{table*}[t]  
	\centering  
	\fontsize{8.8}{15}\selectfont  
	\caption{Different backbone choices of TOAN. All results are with $95\%$ confidence intervals where reported.} \label{Backbone}
	\begin{tabular}{|c|c|c|c|c|c|c|c|c|}  
		\hline
		\multirow{2}{*}{Methods} & 	\multicolumn{2}{c|}{CUB}&\multicolumn{2}{c|}{CARS }&\multicolumn{2}{c|}{DOGS}&\multicolumn{2}{c|}{NABirds}  \cr\cline{2-9}
		&1-shot & 5-shot & 1-shot & 5-shot&1-shot & 5-shot & 1-shot & 5-shot \\
		\hline  \hline
		ConvNet-64~\cite{Sung_2018_CVPR} & {65.34$\pm$0.75}  & {80.43$\pm$0.60} & {65.90$\pm$0.72} & {84.24$\pm$0.48} & {49.30$\pm$0.77} & {67.16$\pm$0.49}  & {70.02$\pm$0.80} & {85.52$\pm$0.50}\\
		ConvNet-128 & 64.56$\pm$0.78  & 80.02$\pm$0.59  & 69.20$\pm$0.72 & 86.39$\pm$0.44  & 50.26$\pm$0.77  & 66.96$\pm$0.66  & 70.90$\pm$0.77  & 85.63$\pm$0.49 \\
		ConvNet-256 & 66.16$\pm$0.80  & 80.72$\pm$0.58  & 68.89$\pm$0.74 & 85.29$\pm$0.46  & 49.68$\pm$0.75  & 67.52$\pm$0.66  & 71.26$\pm$0.76  & 86.42$\pm$0.47 \\
		ConvNet-512~\cite{Gidaris_2019_ICCV} & 66.44$\pm$0.77  & 81.46$\pm$0.54  & 69.59$\pm$0.73 & 86.27$\pm$0.45  & 49.20$\pm$0.74  & 66.75$\pm$0.66  & 72.74$\pm$0.76  & 86.91$\pm$0.50 \\ \hline
		ResNet-64 & 69.25$\pm$0.81  & 81.90$\pm$0.61  & 74.64$\pm$0.76 & 90.20$\pm$0.41  & 53.33$\pm$0.82  & 69.96$\pm$0.70  & 75.98$\pm$0.72  & 89.55$\pm$0.44 \\ 
		ResNet-128 & 68.95$\pm$0.78  & 83.40$\pm$0.58  & 75.14$\pm$0.72 & 90.95$\pm$0.36  & 52.69$\pm$0.81 & 69.95$\pm$0.71  & 76.14$\pm$0.75  & 90.51$\pm$0.38 \\
		ResNet-256~\cite{li2019DN4} & 67.17$\pm$0.81  & 82.09$\pm$0.56  & 76.62$\pm$0.70 & 89.57$\pm$0.40  & 51.83$\pm$0.80  & 69.83$\pm$0.66  & 76.14$\pm$0.75  & 90.21$\pm$0.40 \\
		ResNet-512 & 66.10$\pm$0.86  & 82.27$\pm$0.60  & 75.28$\pm$0.72 & 87.45$\pm$0.48  & 49.77$\pm$0.86  & 69.29$\pm$0.70  & 76.24$\pm$0.77  & 89.88$\pm$0.43 \\ \hline
	\end{tabular}  
\end{table*}

\begin{table*}[!hbpt]  
	\centering  
 	\fontsize{8.8}{15.0}\selectfont  
	\caption{Investigation of model complexity. Model size indicates the number of parameters for each model, and the Inference Time is the testing time for each input query image.} \label{TIME}
	\begin{tabular}{|c|c|c|c|c|c|}  
		\hline
		\multirow{2}{*}{Methods} & 	\multicolumn{5}{c|}{CUB data set} \cr\cline{2-6}
		&1-shot (\%) & 5-shot (\%) & Model Size & Inference Time ($10^{-3}$ s) & Feature Dim \\
		\hline
		ProtoNet~\cite{snell2017prototypical} & 53.88 & 70.85 & 113,088 & 0.69 & 64 \\
	    MatchingNet~\cite{vinyals2016matching} & 57.59 & 70.57 & 113,088 & 0.68 & 64 \\
	    RelationNet~\cite{Sung_2018_CVPR} & 59.82 & 71.83 & 228,686 & 1.14 & 128 \\
	    DN4~\cite{li2019DN4} & 55.60 & 77.64 & 112,832 & 15.20 & 64 \\
	    PABN$_{cpt}$~\cite{huang2019compare}  & 63.36 & 74.71 & 375,361 & 8.65& 4096 \\ 
	    LRPABN$_{cpt}$~\cite{huang2019low}  & 63.63 & 76.06 & 344,251 & 2.53 & 512 \\ \hline
	    TOAN  & 65.60  & 78.93 & 198,417 & 0.66 & 64 \\
	    TOAN  & 65.61  & 79.81 & 237,585 & 0.87 & 128 \\
	    TOAN  & 64.69  & 80.35 & 315,921 & 1.04 & 256 \\
	    TOAN  & 64.17  & 79.19 & 472,593 & 1.23 & 512 \\
	    TOAN  & 65.34  & 80.43 & 785,937 & 2.34 & 1024 \\
	   \hline
	\end{tabular}  
\end{table*}

\begin{table*}[!hbpt]  
	\centering  
 	\fontsize{8.8}{15.0}\selectfont  
	\caption{Investigation of model scalability. Model size indicates the number of parameters for each model.} \label{10W}
	\begin{tabular}{|c|c|c|c|c|c|}  
		\hline
		\multirow{2}{*}{Methods} & 	\multicolumn{5}{c|}{CUB data set} \cr\cline{2-6}
		&10-way-1-shot (\%) & 10-way-5-shot (\%) & Model Size & Inference Time ($10^{-3}$ s) & Feature Dim \\
		\hline
		ProtoNet~\cite{snell2017prototypical} & 37.50$\pm$0.48 & 57.46$\pm$0.56 & 113,088 & 0.74 & 64 \\
	    MatchingNet~\cite{vinyals2016matching} & 40.85$\pm$0.50 & 58.07$\pm$0.55 & 113,088 & 0.76 & 64 \\
	    RelationNet~\cite{Sung_2018_CVPR} & 42.69$\pm$0.52 & 59.37$\pm$0.58 & 228,686 & 1.52 & 128 \\
	    TOAN  & 50.95$\pm$0.57  & 68.82$\pm$0.57 & 198,417 & 1.05 & 64 \\
	   \hline
	\end{tabular}  
\end{table*}

\begin{figure*}[t]
	\centering
	\subfloat[TOAN, 80.67\% accuracy.] {\includegraphics[width=0.4\textwidth]{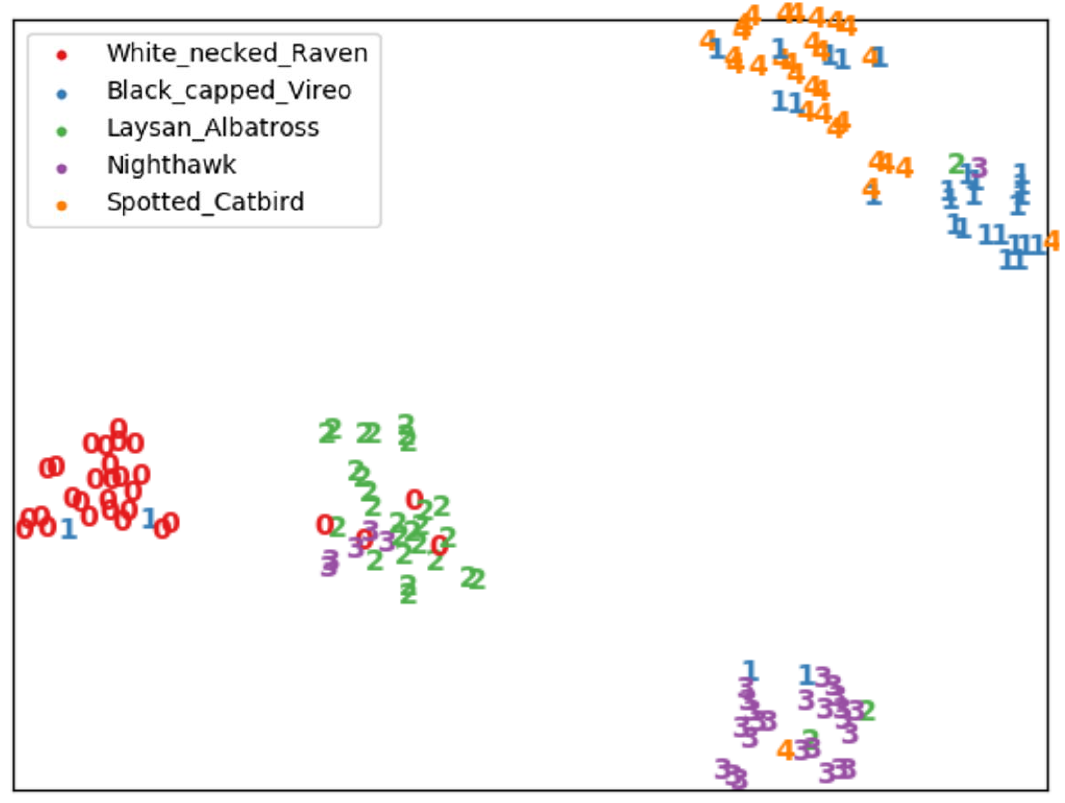} \label{sub01}} 
	\subfloat[RelationNet, 76.67\% accuracy.] {\includegraphics[width=0.4\textwidth]{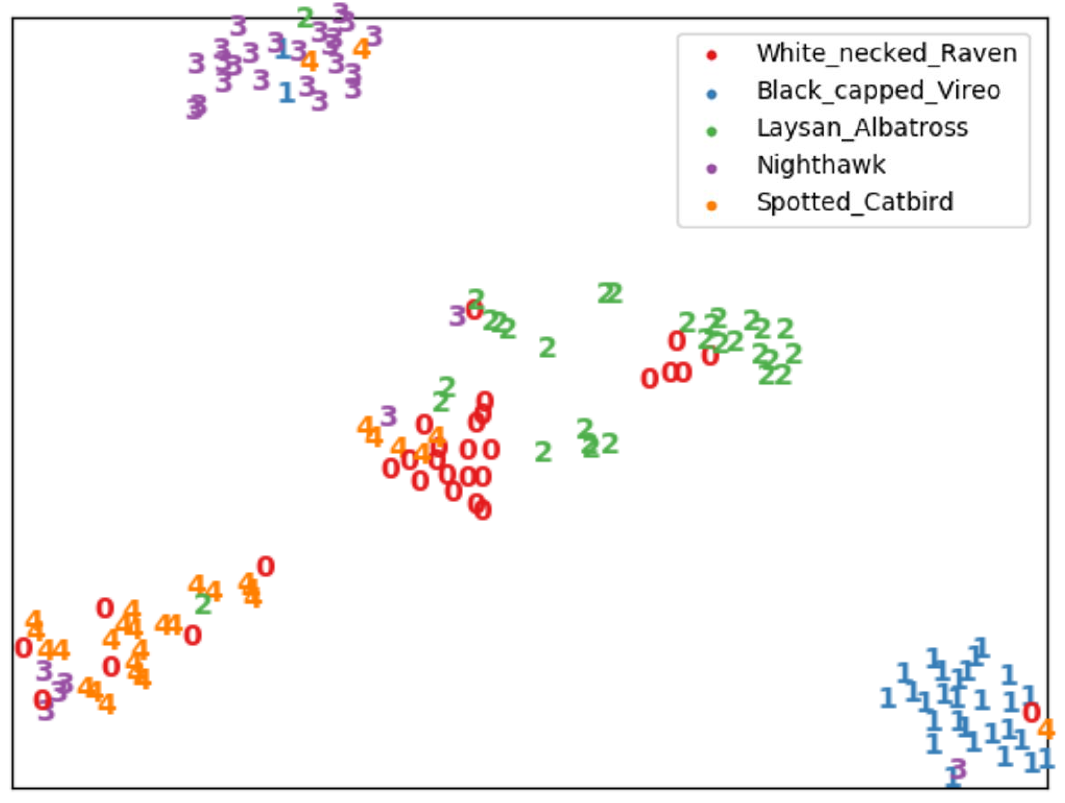} \label{sub02}}
	\caption{t-SNE visualization of the features learned by TOAN, five classes are randomly selected, and in each class, 30 query images are randomly chosen.
	\textcolor{black}{Different colored numbers are used to denote different classes, \textit{i.e.}, red zero represents the white-necked raven, blue one denotes the blacked capped vireo, green two represents the Laysan albatross, purple three denotes the nighthawk, and yellow four denotes the spotted catbird. Moreover, each sample is represented by its corresponding number.}  
	\ref{sub01} and \ref{sub02} show the t-SNE results of TOAN and RelationNet, respectively.}
	\label{TSNE}
\end{figure*}

\paragraph{Comparison against existing FGFS methods}
The comparisons between TOAN and other state-of-the-art FGFS methods are shown in the upper part of Table~\ref{ALL}.
We conclude that our method compares favorably over existing FGFS approaches.
Specifically, under the 5-way-1-shot setting, the classification accuracies are 65.34\% vs. 63.95\%~\cite{zhang2019power}, 65.90\% vs. 29.63\%~\cite{wei2019piecewise}, 49.30\% vs. 49.10\%~\cite{li2019CovaMNet}, and  70.02\% vs. 67.73\%~\cite{huang2019low} on CUB, CARS, DOGS, and NABirds, respectively. 
Moreover, by replacing the ConvNet-64 with the deeper ResNet-256 model~\cite{li2019DN4}, the accuracy of TOAN:ResNet gets further improvements, \textit{e.g.} under 5-way-5-shot setting, TOAN:ResNet achieves 82.09\%, 89.57\%, 69.83\%, and 90.21\% compared with 80.43\%, 84.24\%, 67.16\%, and 85.52\% of TOAN model on four datasets.

\paragraph{Comparison against Generic FS}
As our experiments were conducted under the few-shot setting, we also investigate how generic few-shot learning methods perform for the fine-grained classification. We report the results of representative generic FS models in the lower part of Table~\ref{ALL}. \textcolor{black}{It can be observed that the proposed TOAN models outperform most of these methods by large margins, which is expected since our models are designed to address both intra- and inter-class variance issues in the fine-grained classification.}

\textcolor{black}{More specifically, compared with DN4~\cite{li2019DN4}, the proposed TOAN achieves the highest accuracy on CUB, DOGS, and NABirds datasets, and achieves the highest performance under the 5-way-1-shot setting on the CARS dataset. DN4 achieves the best performance under the 5-way-5-shot setting on the CARS dataset, and our TOAN model ranks the second best.
In general, the proposed TOAN method holds obvious advantages against DN4 for one-shot tasks and most five-shot tasks.}
{It is worth noting that DN4 employs a deep nearest neighbor network to search the optimal local features in the support set as the class prototypes for a given query image. For the query features, DN4 selects the top $k$-nearest local features in the whole support set based on the cosine similarities between the local features of query and support images. Therefore, with more images from support classes (under 5-shot setting on CARS), DN4 tends to generate relatively accurate prototypes.} 

\textcolor{black}{In the fine-grained classification, different categories share similar appearances. The similarities between any two samples are always high, which means the top $k$-nearest local features sorted by DN4 in different support classes are also similar. This leads to the degeneracy problem for DN4 in dealing with fine-grained classification.  
However, the proposed TOAN aligns the support-query pairs by TOMM. Thus the corresponding positions between two samples achieve the highest spatial similarity. Then a GPBP module is adopted to compare the nuanced differences between the pairs using high-order feature extraction. Therefore, TOAN can learn a more robust representation and generally achieves better performances than DN4.
}

\subsection{Ablation Studies}
\textit{Target-Oriented Matching Mechanism (TOMM):}
First of all, we investigate the effectiveness of TOMM for FGFS tasks. As Table~\ref{TOMM} shows, there is an approximate 5\% averagely increase after adopting TOMM in three FS baselines, among which, when incorporating TOMM into the RelationNet, the model achieves superior performances over other compared approaches. For instance, under the 5-way-5-shot setting, the accuracy of RelationNet+TOMM is 79.75\% vs. 78.79\%~\cite{zhang2019power}, 65.23\% vs. 63.04\%~\cite{li2019CovaMNet}, and  85.01\% vs. 83.38\%~\cite{li2019DN4} on CUB, DOGS, and NABirds.
This verifies that the significant intra-class variance is a crucial issue in FGFS tasks, and the TOMM is an effective mechanism to tackle such problems. 
\textcolor{black}{Furthermore, since the proposed task-agnostic transformation $d(\cdot)$ can better capture the similarities of input pairs, TOAN outperforms TOAN-$w/o~d(\cdot)$, as shown in Table~\ref{Other}. To fully investigate the influence of the output channel size ($c'$) of $d(\cdot)$ ($d_{\alpha}$ and $d_{\beta}$) in Eq. (\ref{TOMM-1}), we employ ResNet-256 \cite{li2019DN4,li2019CovaMNet} as the backbone and experiment with different output channel sizes of the TOMM (the input channel size is fixed as 256). As is reported in Table \ref{channel}, the larger output channel size of $d(\cdot)$ generally achieves a better performance.}

In Fig. \ref{TOMMSOW}, we give the visualization of the TOMM. We utilized the original images to get vivid descriptions of the proposed feature alignment. More specifically, we first resized the original images to the same size as the target-oriented attention map (19$\times$19). Then we multiplied the image with the corresponding attention map to generate the aligned features as Equation~(\ref{TOMM-2}). We observe that for the support image (each row in Fig.~\ref{TOMMSOW}), TOMM transforms its features to match each query (top column images). For instance, in the fourth column in Fig.~\ref{sub002}, the postures of five support cars are reshaped to the same posture of the red query car in the top row.

\textit{Group Pair-wise Bilinear Pooling (GPBP):}
\textcolor{black}{The distance-based frameworks such as MatchingNet~\cite{vinyals2016matching} and ProtoNet~\cite{snell2017prototypical} use the $l_2$ or cosine distance of support-query pairs to conduct the classification, while GPBP aims to learn the distance of support-query pairs with a convolutional network, a classifier is then applied to generate the relation confidences, which is the reason why it is not compatible with MatchingNet or ProtoNet. 
On the other hand, RelationNet~\cite{Sung_2018_CVPR} proposes to use a comparator network to classify queries according to the distance of support-query pairs. Therefore, we combine GPBP with RelationNet to study its capability.}
RelationNet+GPBP brings certain performance gains over RelationNet, as is shown in Table~\ref{GPBP}.
As is expected, after combining TOMM and GPBP together, the complete model TOAN achieves significant improvements over the ablation models, indicating that the TOMM and GPBP can benefit from each other.
From the second results row in Table~\ref{Other}, the grouping~\cite{zhang2017interleaved} model TOAN-GP* achieves analogous performances as TOAN under the 1-shot setting. However, its performance is lower than TOAN under the 5-shot, which verifies the effectiveness of our grouping operation. 

We conducted two additional experiments to furtherly investigate the hyper-parameters of GPBP.
First of all, we evaluate the semantic grouping in Fig. \ref{sub2}. We observe that when the grouping number is less than or equal to eight, a larger group size generally results in higher performances, \textit{e.g.,} the accuracy reaches the highest (80.69\%) when the grouping size equals eight, under 5-shot settings. This indicates the effectiveness of semantic grouping on boosting the discrimination of the bilinear features. When the grouping number is greater than eight, the performances tend to be stable with small fluctuations.

Next, we conducted the selection of feature dimensions, as shown in Fig.~\ref{sub1}. It is observed that a higher dimension brings a slight improvement under the 5-shot setting. For example, the performance is 78.93\% vs. 80.10\%, when the length of the bilinear feature is 64 vs. 2048 under the 5-way-5-shot setting on CUB, and the model works relatively stable for 1-shot experiments.

\textit{Input Image Size for TOAN:} In high-order-based FG methods~\cite{Cui_2017_CVPR,Li_2018_CVPR,Lin_2015_ICCV}, a higher resolution of the input image usually results in a more fine-grained feature, which is consistent with the results of current FGFS models~\cite{huang2019low,zhang2019power}. Therefore, we conducted experiments to investigate the effects of input resolution for TOAN. 
As can be seen from Table~\ref{Other}, TOAN\_224 and TOAN:ResNet\_224 achieve further improvements with a larger $224 \times 224$ input size compared with the smaller $84\times84$ resolution. 

\textit{Different Backbones for TOAN:}
We selected different backbones to investigate our proposed model. First, we adopted the ConvNet-512~\cite{Gidaris_2019_ICCV} as the embedding network, which is derived from ConvNet-64 by increasing the width across layers to 512 channels, and we further revise ConvNet-64 to ConvNet-128, ConvNet-256. Similarly, we designed the ResNet-64, ResNet-128 and ResNet-512. From Table~\ref{Backbone}, we observe that a wider ConvNet-based backbone can result in higher performances in FSFG classification.
On the other hand, deeper backbones can achieve further improvements compare to shallow ones. For instance, ResNet-64 outperforms ConvNet-512 on both 1-shot and 5-shot experiments. Under the 5-shot setting, TOAN achieves relatively stable performances when the width of ResNet changes. 

\subsection{Feature Visualization}
Fig.~\ref{sub01} visualizes the feature distribution of the learned fine-grained features using t-SNE~\cite{maaten2008visualizing}. The features are generated under the 5-way-5-shot setting on the CUB. We used 30 query images per class. As can be observed, the learned features by our TOAN have more compact and separable clusters than RelationNet (Fig.~\ref{sub02}). 

\subsection{Model Complexity and Scalability}
The main complexity of our model is the TOMM operation, which has $O((hw)^2)$, where $h \times w$ represents the size of the convolutional map. In our implementation, $h=w=19$. In general, a deeper convolutional network usually results in a smaller feature map before feeding to the classifier. Therefore, the TOMM operation is more efficient with deeper backbones. We conducted additional experiments to investigate the model size and inference time of TOAN compared with previous works~\cite{snell2017prototypical,vinyals2016matching,Sung_2018_CVPR,li2019DN4,huang2019compare,huang2019low}. As is shown in Table~\ref{TIME}, using the same feature dimension, the proposed TOAN model achieves the best performance compared with other models with a small model size as well as a short time. While using a larger dimension, the classification performance can be further improved.

Most current models~\cite{wei2019piecewise,zhang2019power,huang2019low,li2019CovaMNet} are based on a relatively smaller size of the category (five-way) when dealing with FGFS. To further investigate the scalability of the proposed TOAN, we conducted a larger number of categories experiments, which is referred to \cite{chen2019closer,liu2019fewTPN}. As is shown in Table~\ref{10W}, with the same feature dimension, the proposed TOAN outperforms other baseline models with the comparable model size and inference speed.

\section{Conclusion}\label{con}
In this paper, we have proposed a target-oriented alignment network (TOAN) 
for fine-grained image categorization with limited labeled samples. Specifically, a target-oriented matching mechanism is proposed to eliminate the biases brought by the intra-class variance in fine-grained datasets, which is a crucial issue but less considered in current studies. Moreover, the group pair-wise bilinear pooling is adopted to learn compositional bilinear features. 
We have validated the effectiveness of the proposed model on four benchmark datasets, which achieves the state-of-the-art performance.

\section*{Acknowledgment}

The authors greatly appreciate the financial support from the Rail Manufacturing Cooperative Research Centre (funded jointly by participating rail organizations and the Australian Federal Government’s Business-Cooperative Research Centres Program) through Project R3.7.3 - Rail infrastructure defect detection through video analytics.

\ifCLASSOPTIONcaptionsoff
  \newpage
\fi

\bibliographystyle{IEEEtran}
\bibliography{ieeeTrans}

\begin{thebibliography}{10}
\providecommand{\url}[1]{#1}
\csname url@samestyle\endcsname
\providecommand{\newblock}{\relax}
\providecommand{\bibinfo}[2]{#2}
\providecommand{\BIBentrySTDinterwordspacing}{\spaceskip=0pt\relax}
\providecommand{\BIBentryALTinterwordstretchfactor}{4}
\providecommand{\BIBentryALTinterwordspacing}{\spaceskip=\fontdimen2\font plus
\BIBentryALTinterwordstretchfactor\fontdimen3\font minus
  \fontdimen4\font\relax}
\providecommand{\BIBforeignlanguage}[2]{{%
\expandafter\ifx\csname l@#1\endcsname\relax
\typeout{** WARNING: IEEEtran.bst: No hyphenation pattern has been}%
\typeout{** loaded for the language `#1'. Using the pattern for}%
\typeout{** the default language instead.}%
\else
\language=\csname l@#1\endcsname
\fi
#2}}
\providecommand{\BIBdecl}{\relax}
\BIBdecl

\bibitem{KhoslaYaoJayadevaprakashFeiFei_FGVC2011}
A.~Khosla, N.~Jayadevaprakash, B.~Yao, and F.-F. Li, ``Novel dataset for
  fine-grained image categorization: Stanford dogs,'' in \emph{CVPR Workshop},
  vol.~2, no.~1, 2011.

\bibitem{Horn_2015_CVPR}
G.~V. Horn, S.~Branson, R.~Farrell, S.~Haber, J.~Barry, P.~Ipeirotis,
  P.~Perona, and S.~J. Belongie, ``Building a bird recognition app and large
  scale dataset with citizen scientists: The fine print in fine-grained dataset
  collection,'' in \emph{CVPR}.\hskip 1em plus 0.5em minus 0.4em\relax {IEEE}
  Computer Society, 2015, pp. 595--604.

\bibitem{WahCUB_200_2011}
C.~Wah, S.~Branson, P.~Welinder, P.~Perona, and S.~Belongie, ``{The
  Caltech-UCSD Birds-200-2011 Dataset},'' California Institute of Technology,
  Tech. Rep. CNS-TR-2011-001, 2011.

\bibitem{KrauseStarkDengFei-Fei_3DRR2013}
J.~Krause, M.~Stark, J.~Deng, and L.~Fei{-}Fei, ``3d object representations for
  fine-grained categorization,'' in \emph{ICCV}.\hskip 1em plus 0.5em minus
  0.4em\relax {IEEE} Computer Society, 2013, pp. 554--561.

\bibitem{Cui_2017_CVPR}
Y.~Cui, F.~Zhou, J.~Wang, X.~Liu, Y.~Lin, and S.~J. Belongie, ``Kernel pooling
  for convolutional neural networks,'' in \emph{CVPR}.\hskip 1em plus 0.5em
  minus 0.4em\relax {IEEE} Computer Society, 2017, pp. 3049--3058.

\bibitem{Fu_2017_CVPR}
J.~Fu, H.~Zheng, and T.~Mei, ``Look closer to see better: Recurrent attention
  convolutional neural network for fine-grained image recognition,'' in
  \emph{CVPR}.\hskip 1em plus 0.5em minus 0.4em\relax {IEEE} Computer Society,
  2017, pp. 4476--4484.

\bibitem{Krause_2015_CVPR}
J.~Krause, H.~Jin, J.~Yang, and F.~Li, ``Fine-grained recognition without part
  annotations,'' in \emph{CVPR}.\hskip 1em plus 0.5em minus 0.4em\relax {IEEE}
  Computer Society, 2015, pp. 5546--5555.

\bibitem{Li_2018_CVPR}
P.~Li, J.~Xie, Q.~Wang, and Z.~Gao, ``Towards faster training of global
  covariance pooling networks by iterative matrix square root normalization,''
  in \emph{CVPR}.\hskip 1em plus 0.5em minus 0.4em\relax {IEEE} Computer
  Society, 2018, pp. 947--955.

\bibitem{Lin_2015_ICCV}
T.~Lin, A.~RoyChowdhury, and S.~Maji, ``Bilinear {CNN} models for fine-grained
  visual recognition,'' in \emph{ICCV}.\hskip 1em plus 0.5em minus 0.4em\relax
  {IEEE} Computer Society, 2015, pp. 1449--1457.

\bibitem{Yu_2018_ECCV}
C.~Yu, X.~Zhao, Q.~Zheng, P.~Zhang, and X.~You, ``Hierarchical bilinear pooling
  for fine-grained visual recognition,'' in \emph{ECCV}, ser. Lecture Notes in
  Computer Science, vol. 11220.\hskip 1em plus 0.5em minus 0.4em\relax
  Springer, 2018, pp. 595--610.

\bibitem{zhang2014part}
N.~Zhang, J.~Donahue, R.~B. Girshick, and T.~Darrell, ``Part-based r-cnns for
  fine-grained category detection,'' in \emph{ECCV}, ser. Lecture Notes in
  Computer Science, vol. 8689.\hskip 1em plus 0.5em minus 0.4em\relax Springer,
  2014, pp. 834--849.

\bibitem{zhang2018fine}
Y.~Zhang, H.~Tang, and K.~Jia, ``Fine-grained visual categorization using
  meta-learning optimization with sample selection of auxiliary data,'' in
  \emph{ECCV}, vol. 11212.\hskip 1em plus 0.5em minus 0.4em\relax Springer,
  2018, pp. 241--256.

\bibitem{zheng2019learning}
H.~Zheng, J.~Fu, Z.~Zha, and J.~Luo, ``Learning deep bilinear transformation
  for fine-grained image representation,'' in \emph{NIPS}, 2019, pp.
  4279--4288.

\bibitem{wang2019deep}
Y.~Wang, Q.~Hu, P.~Zhu, L.~Li, B.~Lu, J.~M. Garibaldi, and X.~Li, ``Deep fuzzy
  tree for large-scale hierarchical visual classification,'' \emph{IEEE
  Transactions on Fuzzy Systems}, vol.~28, no.~7, pp. 1395--1406, 2019.

\bibitem{Han}
J.~{Han}, X.~{Yao}, G.~{Cheng}, X.~{Feng}, and D.~{Xu}, ``P-cnn: Part-based
  convolutional neural networks for fine-grained visual categorization,''
  \emph{IEEE Transactions on Pattern Analysis and Machine Intelligence}, pp.
  1--1, 2019.

\bibitem{li2019CovaMNet}
W.~Li, J.~Xu, J.~Huo, L.~Wang, Y.~Gao, and J.~Luo, ``Distribution consistency
  based covariance metric networks for few-shot learning,'' in
  \emph{AAAI}.\hskip 1em plus 0.5em minus 0.4em\relax {AAAI} Press, 2019, pp.
  8642--8649.

\bibitem{wei2019piecewise}
X.-S. Wei, P.~Wang, L.~Liu, C.~Shen, and J.~Wu, ``Piecewise classifier
  mappings: Learning fine-grained learners for novel categories with few
  examples,'' \emph{IEEE TIP}, vol.~28, no.~12, pp. 6116--6125, 2019.

\bibitem{wertheimer2019few}
D.~Wertheimer and B.~Hariharan, ``Few-shot learning with localization in
  realistic settings,'' in \emph{CVPR}.\hskip 1em plus 0.5em minus 0.4em\relax
  Computer Vision Foundation / {IEEE}, 2019, pp. 6558--6567.

\bibitem{zhang2019power}
H.~Zhang and P.~Koniusz, ``Power normalizing second-order similarity network
  for few-shot learning,'' in \emph{WACV}.\hskip 1em plus 0.5em minus
  0.4em\relax {IEEE}, 2019, pp. 1185--1193.

\bibitem{huang2019low}
H.~{Huang}, J.~{Zhang}, J.~{Zhang}, J.~{Xu}, and Q.~{Wu}, ``Low-rank pairwise
  alignment bilinear network for few-shot fine-grained image classification,''
  \emph{IEEE Transactions on Multimedia}, pp. 1--1, 2020.

\bibitem{tmp1}
D.~Lin, X.~Shen, C.~Lu, and J.~Jia, ``Deep lac: Deep localization, alignment
  and classification for fine-grained recognition,'' in \emph{CVPR}, 2015, pp.
  1666--1674.

\bibitem{tmp2}
R.~Farrell, O.~Oza, N.~Zhang, V.~I. Morariu, T.~Darrell, and L.~S. Davis,
  ``Birdlets: Subordinate categorization using volumetric primitives and
  pose-normalized appearance,'' in \emph{ICCV}.\hskip 1em plus 0.5em minus
  0.4em\relax IEEE, 2011, pp. 161--168.

\bibitem{tmp3}
S.~Yang, L.~Bo, J.~Wang, and L.~G. Shapiro, ``Unsupervised template learning
  for fine-grained object recognition,'' in \emph{NIPS}, 2012, pp. 3122--3130.

\bibitem{tmp4}
B.~Yao, A.~Khosla, and L.~Fei-Fei, ``Combining randomization and discrimination
  for fine-grained image categorization,'' in \emph{CVPR}.\hskip 1em plus 0.5em
  minus 0.4em\relax IEEE, 2011, pp. 1577--1584.

\bibitem{vaswani2017attention}
A.~Vaswani, N.~Shazeer, N.~Parmar, J.~Uszkoreit, L.~Jones, A.~N. Gomez,
  L.~Kaiser, and I.~Polosukhin, ``Attention is all you need,'' in \emph{NIPS},
  2017, pp. 5998--6008.

\bibitem{hou2019cross}
R.~Hou, H.~Chang, B.~Ma, S.~Shan, and X.~Chen, ``Cross attention network for
  few-shot classification,'' in \emph{NIPS}, 2019, pp. 4005--4016.

\bibitem{biederman1987recognition}
I.~Biederman, ``Recognition-by-components: a theory of human image
  understanding.'' \emph{Psychological review}, vol.~94, no.~2, p. 115, 1987.

\bibitem{hoffman1984parts}
D.~D. Hoffman and W.~A. Richards, ``Parts of recognition,'' \emph{Cognition},
  vol.~18, no. 1-3, pp. 65--96, 1984.

\bibitem{marr1978representation}
D.~Marr and H.~K. Nishihara, ``Representation and recognition of the spatial
  organization of three-dimensional shapes,'' \emph{Proceedings of the Royal
  Society of London. Series B. Biological Sciences}, vol. 200, no. 1140, pp.
  269--294, 1978.

\bibitem{simon2015neural}
M.~Simon and E.~Rodner, ``Neural activation constellations: Unsupervised part
  model discovery with convolutional networks,'' in \emph{ICCV}.\hskip 1em plus
  0.5em minus 0.4em\relax {IEEE} Computer Society, 2015, pp. 1143--1151.

\bibitem{zhang2016picking}
X.~Zhang, H.~Xiong, W.~Zhou, W.~Lin, and Q.~Tian, ``Picking deep filter
  responses for fine-grained image recognition,'' in \emph{CVPR}.\hskip 1em
  plus 0.5em minus 0.4em\relax {IEEE} Computer Society, 2016, pp. 1134--1142.

\bibitem{zheng2017learning}
H.~Zheng, J.~Fu, T.~Mei, and J.~Luo, ``Learning multi-attention convolutional
  neural network for fine-grained image recognition,'' in \emph{ICCV}.\hskip
  1em plus 0.5em minus 0.4em\relax {IEEE} Computer Society, 2017, pp.
  5219--5227.

\bibitem{hu2019weakly}
P.~Hu, X.~Sun, K.~Saenko, and S.~Sclaroff, ``Weakly-supervised compositional
  feature aggregation for few-shot recognition,'' \emph{arXiv preprint
  arXiv:1906.04833}, 2019.

\bibitem{zhang2017interleaved}
T.~Zhang, G.~Qi, B.~Xiao, and J.~Wang, ``Interleaved group convolutions,'' in
  \emph{ICCV}.\hskip 1em plus 0.5em minus 0.4em\relax {IEEE} Computer Society,
  2017, pp. 4383--4392.

\bibitem{Chen_2019_CVPR}
Y.~Chen, Y.~Bai, W.~Zhang, and T.~Mei, ``Destruction and construction learning
  for fine-grained image recognition,'' in \emph{CVPR}.\hskip 1em plus 0.5em
  minus 0.4em\relax Computer Vision Foundation / {IEEE}, 2019, pp. 5157--5166.

\bibitem{engin2018deepkspd}
M.~Engin, L.~Wang, L.~Zhou, and X.~Liu, ``Deepkspd: Learning
  kernel-matrix-based {SPD} representation for fine-grained image
  recognition,'' in \emph{ECCV}, ser. Lecture Notes in Computer Science, vol.
  11206.\hskip 1em plus 0.5em minus 0.4em\relax Springer, 2018, pp. 629--645.

\bibitem{Ge_2019_CVPR}
W.~Ge, X.~Lin, and Y.~Yu, ``Weakly supervised complementary parts models for
  fine-grained image classification from the bottom up,'' in \emph{CVPR}.\hskip
  1em plus 0.5em minus 0.4em\relax Computer Vision Foundation / {IEEE}, 2019,
  pp. 3034--3043.

\bibitem{zheng2019looking}
H.~Zheng, J.~Fu, Z.~Zha, and J.~Luo, ``Looking for the devil in the details:
  Learning trilinear attention sampling network for fine-grained image
  recognition,'' in \emph{CVPR}.\hskip 1em plus 0.5em minus 0.4em\relax
  Computer Vision Foundation / {IEEE}, 2019, pp. 5012--5021.

\bibitem{he2019fine}
X.~He and Y.~Peng, ``Fine-grained visual-textual representation learning,''
  \emph{{IEEE} Trans. Circuits Syst. Video Techn.}, vol.~30, no.~2, pp.
  520--531, 2020.

\bibitem{cai2017higher}
S.~Cai, W.~Zuo, and L.~Zhang, ``Higher-order integration of hierarchical
  convolutional activations for fine-grained visual categorization,'' in
  \emph{ICCV}.\hskip 1em plus 0.5em minus 0.4em\relax {IEEE} Computer Society,
  2017, pp. 511--520.

\bibitem{koniusz2018deeper}
P.~Koniusz, H.~Zhang, and F.~Porikli, ``A deeper look at power
  normalizations,'' in \emph{CVPR}.\hskip 1em plus 0.5em minus 0.4em\relax
  {IEEE} Computer Society, 2018, pp. 5774--5783.

\bibitem{hu2020attentional}
Y.~Hu, Y.~Yang, J.~Zhang, X.~Cao, and X.~Zhen, ``Attentional kernel encoding
  networks for fine-grained visual categorization,'' \emph{IEEE Transactions on
  Circuits and Systems for Video Technology}, pp. 1--1, 2020.

\bibitem{Gao_2016_CVPR}
Y.~Gao, O.~Beijbom, N.~Zhang, and T.~Darrell, ``Compact bilinear pooling,'' in
  \emph{CVPR}.\hskip 1em plus 0.5em minus 0.4em\relax {IEEE} Computer Society,
  2016, pp. 317--326.

\bibitem{Kong_2017_CVPR}
S.~Kong and C.~C. Fowlkes, ``Low-rank bilinear pooling for fine-grained
  classification,'' in \emph{CVPR}.\hskip 1em plus 0.5em minus 0.4em\relax
  {IEEE} Computer Society, 2017, pp. 7025--7034.

\bibitem{kim2016hadamard}
J.~Kim, K.~W. On, W.~Lim, J.~Kim, J.~Ha, and B.~Zhang, ``Hadamard product for
  low-rank bilinear pooling,'' in \emph{ICLR}.\hskip 1em plus 0.5em minus
  0.4em\relax OpenReview.net, 2017.

\bibitem{huang2019compare}
H.~Huang, J.~Zhang, J.~Zhang, Q.~Wu, and J.~Xu, ``Compare more nuanced:
  Pairwise alignment bilinear network for few-shot fine-grained learning,'' in
  \emph{ICME}.\hskip 1em plus 0.5em minus 0.4em\relax {IEEE}, 2019, pp. 91--96.

\bibitem{pahde2018discriminative}
F.~Pahde, M.~Nabi, T.~Klein, and P.~J{\"{a}}hnichen, ``Discriminative
  hallucination for multi-modal few-shot learning,'' in \emph{ICIP}.\hskip 1em
  plus 0.5em minus 0.4em\relax {IEEE}, 2018, pp. 156--160.

\bibitem{tsutsui2019meta}
S.~Tsutsui, Y.~Fu, and D.~J. Crandall, ``Meta-reinforced synthetic data for
  one-shot fine-grained visual recognition,'' in \emph{NIPS}, 2019, pp.
  3057--3066.

\bibitem{he2018only}
X.~He and Y.~Peng, ``Only learn one sample: Fine-grained visual categorization
  with one sample training,'' in \emph{ACM MM}.\hskip 1em plus 0.5em minus
  0.4em\relax {ACM}, 2018, pp. 1372--1380.

\bibitem{zhumulti}
Y.~Zhu, C.~Liu, and S.~Jiang, ``Multi-attention meta learning for few-shot
  fine-grained image recognition,'' in \emph{AAAI}, 2020, pp. 1090--1096.

\bibitem{CVPRWorkshop}
B.~Haney and A.~Lavin, ``Fine-grain few-shot vision via domain knowledge as
  hyperspherical priors,'' \emph{CoRR}, vol. abs/2005.11450, 2020.

\bibitem{CVPRWorkshop2}
P.~Mettes, E.~van~der Pol, and C.~Snoek, ``Hyperspherical prototype networks,''
  in \emph{NIPS}, 2019, pp. 1485--1495.

\bibitem{w6}
``Few-shot fine-grained classification with spatial attentive comparison,''
  \emph{Knowledge-Based Systems}, p. 106840, 2021.

\bibitem{munkhdalai2017meta}
T.~Munkhdalai and H.~Yu, ``Meta networks,'' in \emph{ICML}, ser. Proceedings of
  Machine Learning Research, vol.~70.\hskip 1em plus 0.5em minus 0.4em\relax
  {PMLR}, 2017, pp. 2554--2563.

\bibitem{santoro2016meta}
A.~Santoro, S.~Bartunov, M.~Botvinick, D.~Wierstra, and T.~P. Lillicrap,
  ``Meta-learning with memory-augmented neural networks,'' in \emph{ICML}, ser.
  {JMLR} Workshop and Conference Proceedings, vol.~48.\hskip 1em plus 0.5em
  minus 0.4em\relax JMLR.org, 2016, pp. 1842--1850.

\bibitem{chen2019closer}
W.~Chen, Y.~Liu, Z.~Kira, Y.~F. Wang, and J.~Huang, ``A closer look at few-shot
  classification,'' in \emph{ICLR}.\hskip 1em plus 0.5em minus 0.4em\relax
  OpenReview.net, 2019.

\bibitem{pmlr-v70-finn17a}
C.~Finn, P.~Abbeel, and S.~Levine, ``Model-agnostic meta-learning for fast
  adaptation of deep networks,'' in \emph{ICML}, ser. Proceedings of Machine
  Learning Research, vol.~70.\hskip 1em plus 0.5em minus 0.4em\relax {PMLR},
  2017, pp. 1126--1135.

\bibitem{rajeswaran2019meta}
A.~Rajeswaran, C.~Finn, S.~M. Kakade, and S.~Levine, ``Meta-learning with
  implicit gradients,'' in \emph{NIPS}, 2019, pp. 113--124.

\bibitem{Sachin2017}
S.~Ravi and H.~Larochelle, ``Optimization as a model for few-shot learning,''
  in \emph{ICLR}.\hskip 1em plus 0.5em minus 0.4em\relax OpenReview.net, 2017.

\bibitem{li2019DN4}
W.~Li, L.~Wang, J.~Xu, J.~Huo, Y.~Gao, and J.~Luo, ``Revisiting local
  descriptor based image-to-class measure for few-shot learning,'' in
  \emph{CVPR}.\hskip 1em plus 0.5em minus 0.4em\relax Computer Vision
  Foundation / {IEEE}, 2019, pp. 7260--7268.

\bibitem{snell2017prototypical}
J.~Snell, K.~Swersky, and R.~S. Zemel, ``Prototypical networks for few-shot
  learning,'' in \emph{NIPS}, 2017, pp. 4077--4087.

\bibitem{Sung_2018_CVPR}
F.~Sung, Y.~Yang, L.~Zhang, T.~Xiang, P.~H.~S. Torr, and T.~M. Hospedales,
  ``Learning to compare: Relation network for few-shot learning,'' in
  \emph{CVPR}.\hskip 1em plus 0.5em minus 0.4em\relax {IEEE} Computer Society,
  2018, pp. 1199--1208.

\bibitem{vinyals2016matching}
O.~Vinyals, C.~Blundell, T.~Lillicrap, K.~Kavukcuoglu, and D.~Wierstra,
  ``Matching networks for one shot learning,'' in \emph{NIPS}, 2016, pp.
  3630--3638.

\bibitem{zhang2019few}
C.~Zhang, C.~Li, and J.~Cheng, ``Few-shot visual classification using image
  pairs with binary transformation,'' \emph{IEEE Transactions on Circuits and
  Systems for Video Technology}, pp. 1--1, 2019.

\bibitem{gidaris2018dynamic}
S.~Gidaris and N.~Komodakis, ``Dynamic few-shot visual learning without
  forgetting,'' in \emph{CVPR}.\hskip 1em plus 0.5em minus 0.4em\relax {IEEE}
  Computer Society, 2018, pp. 4367--4375.

\bibitem{Hao_2019_ICCV}
F.~Hao, F.~He, J.~Cheng, L.~Wang, J.~Cao, and D.~Tao, ``Collect and select:
  Semantic alignment metric learning for few-shot learning,'' in
  \emph{ICCV}.\hskip 1em plus 0.5em minus 0.4em\relax {IEEE}, 2019, pp.
  8459--8468.

\bibitem{wu2019parn}
Z.~Wu, Y.~Li, L.~Guo, and K.~Jia, ``{PARN:} position-aware relation networks
  for few-shot learning,'' in \emph{ICCV}.\hskip 1em plus 0.5em minus
  0.4em\relax {IEEE}, 2019, pp. 6658--6666.

\bibitem{CSVT2020}
W.~{Jiang}, K.~{Huang}, J.~{Geng}, and X.~{Deng}, ``Multi-scale metric learning
  for few-shot learning,'' \emph{IEEE Transactions on Circuits and Systems for
  Video Technology}, pp. 1--1, 2020.

\bibitem{liu2019fewTPN}
Y.~Liu, J.~Lee, M.~Park, S.~Kim, E.~Yang, S.~J. Hwang, and Y.~Yang, ``Learning
  to propagate labels: Transductive propagation network for few-shot
  learning,'' in \emph{ICLR}.\hskip 1em plus 0.5em minus 0.4em\relax
  OpenReview.net, 2019.

\bibitem{w3}
D.~Zhang, J.~Han, L.~Zhao, and T.~Zhao, ``From discriminant to complete:
  Reinforcement searching-agent learning for weakly supervised object
  detection,'' \emph{IEEE Transactions on Neural Networks and Learning
  Systems}, vol.~31, no.~12, pp. 5549--5560, 2020.

\bibitem{w4}
D.~Zhang, J.~Han, G.~Guo, and L.~Zhao, ``Learning object detectors with
  semi-annotated weak labels,'' \emph{IEEE Transactions on Circuits and Systems
  for Video Technology}, vol.~29, no.~12, pp. 3622--3635, 2018.

\bibitem{w1}
M.~Oquab, L.~Bottou, I.~Laptev, and J.~Sivic, ``Is object localization for
  free?-weakly-supervised learning with convolutional neural networks,'' in
  \emph{Proceedings of the IEEE conference on computer vision and pattern
  recognition}, 2015, pp. 685--694.

\bibitem{w2}
J.~Peyre, J.~Sivic, I.~Laptev, and C.~Schmid, ``Weakly-supervised learning of
  visual relations,'' in \emph{Proceedings of the ieee international conference
  on computer vision}, 2017, pp. 5179--5188.

\bibitem{w5}
D.~Zhang, J.~Han, L.~Yang, and D.~Xu, ``Spftn: a joint learning framework for
  localizing and segmenting objects in weakly labeled videos,'' \emph{IEEE
  transactions on pattern analysis and machine intelligence}, vol.~42, no.~2,
  pp. 475--489, 2018.

\bibitem{kingma2015adam}
D.~P. Kingma and J.~Ba, ``Adam: {A} method for stochastic optimization,'' in
  \emph{ICLR}, 2015.

\bibitem{Gidaris_2019_ICCV}
S.~Gidaris, A.~Bursuc, N.~Komodakis, P.~P{\'{e}}rez, and M.~Cord, ``Boosting
  few-shot visual learning with self-supervision,'' in \emph{ICCV}.\hskip 1em
  plus 0.5em minus 0.4em\relax {IEEE}, 2019, pp. 8058--8067.

\bibitem{maaten2008visualizing}
L.~v.~d. Maaten and G.~Hinton, ``Visualizing data using t-sne,'' \emph{JMLR},
  vol.~9, no. Nov, pp. 2579--2605, 2008.

\end{thebibliography}

\end{document}